\documentclass[11pt,a4paper]{article}
\usepackage[hyperref]{acl2021}
\usepackage{times}
\usepackage{latexsym}

\usepackage{microtype}

\aclfinalcopy 
\def\aclpaperid{166} 

\title{Generate, Prune, Select: A Pipeline for Counterspeech Generation \\against Online Hate Speech}

\author{Wanzheng Zhu \and Suma Bhat\\
	University of Illinois at Urbana-Champaign, USA \\
	\texttt{wz6@illinois.edu}, \texttt{spbhat2@illinois.edu}}

\date{}
\usepackage{graphicx, amsmath, amsfonts, amssymb, xspace, color, enumitem}
\usepackage{amsthm}
\usepackage{amsfonts}
\usepackage{xcolor}
\usepackage{booktabs}
\usepackage{tabularx} 
\usepackage[noend,ruled,linesnumbered]{algorithm2e}
\usepackage{multirow}
\usepackage{subcaption}

\newcommand{\ie}{{i.e.}\xspace} 
\newcommand{\eg}{{e.g.}\xspace} 
\newcommand{\our}{\textsc{GPS}\xspace}
\newcommand{\nop}[1]{}

\newcommand{\ppa}{0.9}
\usepackage{comment}

\begin{document}
\maketitle
\begin{abstract}
	\textbf{Warning}: \textit{this paper contains content that may be offensive or upsetting.}
	
	Countermeasures to effectively fight the ever increasing hate speech online without blocking freedom of speech is of great social interest. 
	Natural Language Generation (NLG), is uniquely capable of developing scalable solutions.
	However, off-the-shelf NLG methods are primarily sequence-to-sequence neural models and they are limited in that they generate commonplace, repetitive and safe responses regardless of the hate speech (\eg, ``Please refrain from using such language.") or irrelevant responses, making them ineffective for de-escalating hateful conversations. 
	In this paper, we design a three-module pipeline approach to effectively improve the \textit{diversity} and \textit{relevance}. 
	Our proposed pipeline first generates various counterspeech candidates by a generative model to promote \textit{diversity}, then filters the ungrammatical ones using a BERT model, and finally selects the most \textit{relevant} counterspeech response using a novel retrieval-based method. 
	Extensive Experiments on three representative datasets demonstrate the efficacy of our approach in generating diverse and relevant counterspeech. 
\end{abstract}

\section{Introduction}
\label{sec:introduction}
Hate speech is any form of expression through which speakers intend to vilify, humiliate, or incite hatred against a group or a class of persons on the basis of some characteristics, including race, religion, skin color, sexual identity, gender identity, ethnicity, disability, or national origin \cite{ward1997free,nockleby2000hate}. 
Its ever-growing increase on the Internet makes it a problem of significant societal concern \cite{williams2019hatred}; effective countermeasures call for not blocking freedom of speech by means of censorship or active moderation \cite{gagliardone2015countering,strossen2018hate}. 
A very promising countermeasure is {\em counterspeech}---a response that provides non-negative feedback through fact-bound arguments and broader perspectives to mitigate hate speech and fostering a more harmonious conversation in social platforms \cite{schieb2016governing,munger2017tweetment,mathew2018analyzing,shin2018data}. Counterspeech as a measure to combat abusive language online is also promoted in active campaigns such as ``Get The Trolls Out".\footnote{\url{https://getthetrollsout.org/stoppinghate}}

\begin{table}[t]
	\small
	\centering
	\begin{tabular}{cc}
		\toprule
		\multicolumn{1}{p{0.07\textwidth}}{\textbf{Hate Speech}: }  & \multicolumn{1}{p{0.36\textwidth}}{I am done with Islam and isis. All Muslims should be sent to their homeland. Britain will be better without their violence and ideology.}  \\
		\midrule
		\multicolumn{1}{p{0.07\textwidth}}{\textbf{Expert}: } & \multicolumn{1}{p{0.36\textwidth}}{I agree that ISIS is an evil aberration, but to extend this to include up to 3 million people just in the UK is just plain silly.}  \\
		\midrule
		\multicolumn{1}{p{0.07\textwidth}}{\textbf{Common-place}: } & \multicolumn{1}{p{0.36\textwidth}}{Hate speech is not tolerated. Please review our user policies. 
			Thank you for your cooperation.}  \\
		\midrule
		\multicolumn{1}{p{0.07\textwidth}}{\textbf{Not relevant}: } & \multicolumn{1}{p{0.36\textwidth}}{Use of the r-word is unacceptable  as it demeans and insults people with disabilities.}  \\
		\midrule
	\end{tabular}
	\caption{An illustrative example of hate speech and counterspeech. 
	}
	\label{table:illustrative_example}
\end{table}

What makes an effective counterspeech?
Informed by psychosocial and linguistic studies on counterspeech \cite{mathew2019thou} and the large number of effective counterspeech examples created by crowdsourcing \cite{qian2019benchmark} and by experts \cite{chung2019conan}, 
we identify that effective counterspeech should be \textbf{diverse} and \textbf{relevant} to the hate speech instance. 
\textit{Diversity} is the requirement that a collection of counterspeech should not be largely commonplace, repetitive and safe responses without regard to the target or type of hate speech (\eg, ``Please refrain from using such language."). 
\textit{Relevance} refers to the property that counterspeech should directly address and target the central aspects of the hate speech, enabling coherent conversations rather than irrelevant or off-topic ones (\eg, the hate speech instance targets an ethnic group, while the counterspeech talks about people with disabilities). 
Comparative examples are shown in Table \ref{table:illustrative_example} where we list some counterspeech that lack diversity or relevance.

While NLG systems (in particular, sequence-to-sequence models) offer much promise for generating text at scale \cite{sutskever2014sequence,zhu2018texygen,lewis2020bart}, the quality of the outputs is modest in the context of the requirements identified above. 
Indeed, \citet{qian2019benchmark}, the only existing quality work on counterspeech generation, has highlighted their limitations: the responses are largely commonplace and sometimes irrelevant. 
These limitations apply more broadly to general conversational language generation tasks, arising primarily due to the intrinsic end-to-end training nature of a single sequence-to-sequence architecture \cite{sordoni2015neural,li2016diversity,serban2017hierarchical,jiang2018sequence}. 
Model refinements to account for these limitations have been addressed individually: improved diversity \cite{li2016diversity,xu2018diversity} or improved relevance \cite{gao2019jointly,li2020relevance}. 
However, combining these improvements into a single model is not straightforward. Such is the goal of this paper.

We tackle the problem from an entirely novel angle by proposing a three-module pipeline approach, \textit{Generate}, \textit{Prune}, \textit{Select} (denoted as ``GPS") to ensure the generated sentences adhere to the required properties of diversity and relevance. 
First, the \textit{Candidate Generation} module generates a large number of diverse response candidates using a generative model. 
As such, a large candidate pool is made available for selection, which accounts for improved diversity. 
Second, the \textit{Candidate Pruning} module prunes the ungrammatical candidates from the candidate pool. 
Last, from the pruned counterspeech candidate pool, the \textit{Response Selection} module selects the most relevant counterspeech for a given hate speech instance by a novel retrieval-based response selection method. 

We demonstrate the efficacy of \our, the first pipeline approach for counterspeech generation, by a systematic comparison with other competitive NLG approaches in generating \textit{diverse} and \textit{relevant} counterspeech. 
We derive new state-of-the-art results on three benchmark datasets by showing improved diversity and relevance using both automatic and human evaluations.

\section{Proposed Model}
\label{sec:model}

\begin{figure*}[ht]
	\centering
	\includegraphics[width=0.99\linewidth]{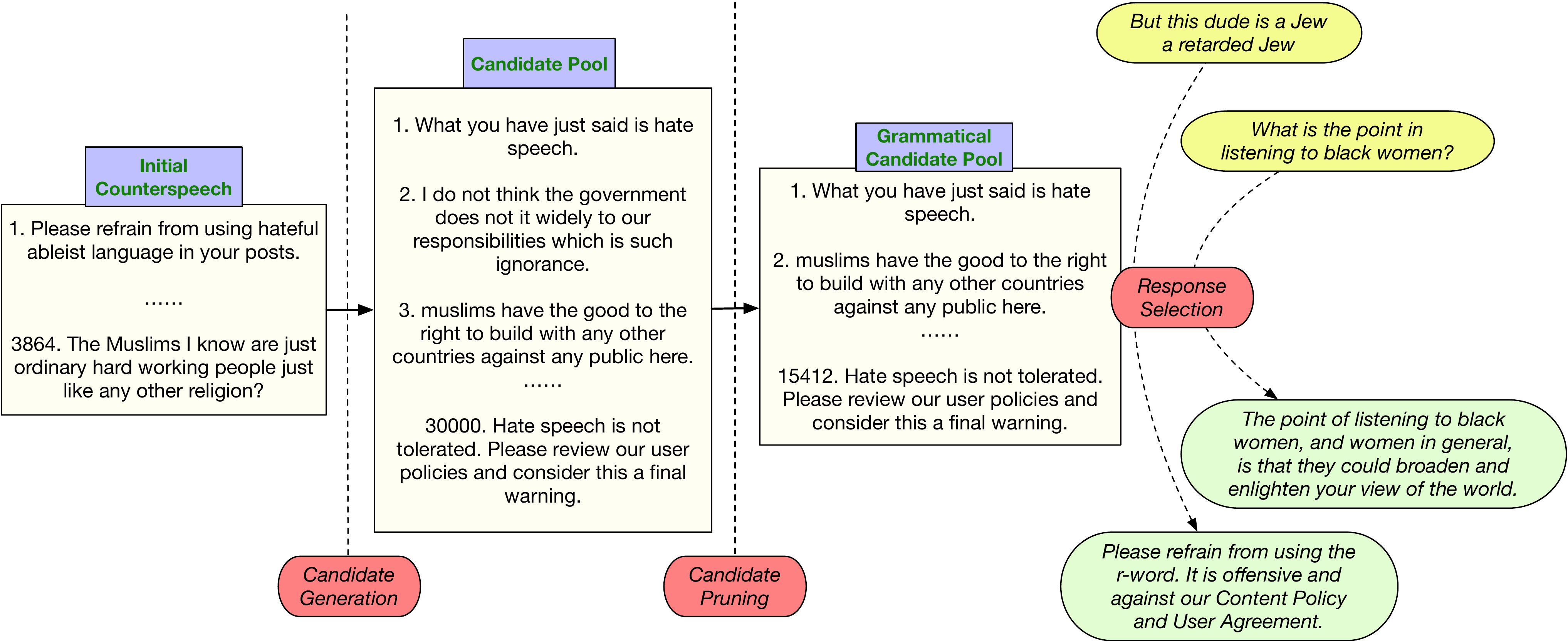}
	\caption{Overview of \our. The red ovals correspond to the individual modules. 
	}
	\label{fig:model}
\end{figure*}

We assume access to a corpus of labeled pairs of conversations $\mathcal{D} = \{(x_1, y_1), (x_2, y_2), ..., (x_n, y_n) \}$, where $x_i$ is a hate speech and $y_i$ is the appropriate counterspeech as decided by experts or by crowdsourcing. 
The goal is to learn a model that takes as input a hate speech $x$ and outputs a counterspeech $y$. 
A motivating example is shown in Table \ref{table:illustrative_example}. 
Most importantly, we aim at generating \textit{diverse} and \textit{relevant} counterspeech. 
We present an overview of the model in Figure \ref{fig:model} and describe each module in detail below.

\subsection{Candidate Generation}
The main goal of this module is to create a diverse candidate pool for counterspeech selection. 
We extract all available counterspeech instances $Y = [y_1, y_2, ..., y_n]$ from the training dataset and enlarge the counterspeech pool by a generative model. 

Specifically, we utilize an RNN-based variational autoencoder  \cite{bowman2016generating}, that incorporates the global distributed latent representations of all sentences to generate candidates. 
Both the encoder and the decoder have two layers with 512 nodes each, and we use two highway network layers \cite{srivastava2015training} to facilitate robust training. 
Like all other generative models, it aims to maximize the lower bound of the likelihood $\mathcal{L}$ of generating the training data $Y$, 
$$\mathcal{L} = -KL(q_{\theta}(z|y) \: || \: p(z)) + \mathbb{E}_{q_{\theta}(z|y)}[\log p_{\theta}(y | z)]$$
\noindent where $\theta$ denotes all parameters of the generative model, $z$ is a latent variable having a Gaussian distribution with a diagonal covariance matrix, $p$ denotes the prior distribution, $q$ denotes the posterior distribution, and $KL$ denotes the KL-divergence \cite{kullback1951information}. 
In the training process, we apply the KL annealing technique \cite{bowman2016generating} to prevent the undesirable stable equilibrium problem (\ie, the first term of the likelihood function $KL(q_{\theta}(z|y) \: || \: p({z}))$ becomes zero). 
Upon the completion of the training, we generate candidates by simply decoding from noise $\epsilon$ sampled from a standard Gaussian distribution (\ie, $\epsilon \sim \mathcal{N}(0, 1)$). 

As demonstrated by \citet{bowman2016generating} (and as inferred from our own experiments described in Section~\ref{sec:exp}), the generative model not only captures holistic properties of sentences such as style, topic, and high-level syntactic features, but also produces \textit{diverse} candidates.

\subsection{Candidate Pruning}
Though candidates generated by such an RNN-based variational autoencoder are diverse, they are not always grammatical as pointed out by \citet{bowman2016generating}. 
Therefore, in this module, we prune the candidate list and retain only the grammatical ones. 
Toward this, we train a grammaticality classifier on the corpus of linguistic acceptability (CoLA) \cite{warstadt2018neural}, a dataset with 10,657 English sentences labeled as grammatical or ungrammatical from  linguistics publications. 
We select BERT \cite{devlin2019bert} as the classification model, and fine-tune it on the CoLA dataset. 
The choice of BERT is to best capture both the syntactic and the contextual information, and we select the `bert-base-cased' model for its better computational efficiency. 

\subsection{Response Selection}
We now have a collection of diverse and grammatical counterspeech responses. Finally, we aim to select the most \textit{relevant} response to a given hate speech instance. 

Taking into consideration the limited training instances that are realistically available \cite{chung2019conan,qian2019benchmark}, 
and inspired by the recent success of pretrained models \cite{devlin2019bert}, we innovate on a pretrained response selection model for task-oriented dialogue systems \cite{henderson2019training} and perform fine-tuning on our dataset. 
\citet{henderson2019training} proposed two response selection methods, but we find that neither of them is well-suited for our task. 
\begin{enumerate}[leftmargin=*]
	\item \textit{Train a response selection classifier with the negative sampling technique}: It relies on randomly drawing other candidates from the candidate pool as negative examples. 
	However, in our task, one hate speech instance usually has multiple appropriate counterspeech instances. 
	For example, given the hate speech in Table \ref{table:illustrative_example}, there are many other instances that can work as quality counterspeech, such as ``You cannot blame all people for the actions of a few. Banning something altogether will not solve anything." or ``Does prohibition of anything ever work? I thought religious tolerance was one of our `British values'?". 
	Therefore, many wrongly chosen negative examples may negatively impact the inductive bias of the response selection classifier. 
	\item \textit{Select by cosine similarity}: we point out that the embeddings of the input (hate speech) and the responses (counterspeech candidates) do not share the same latent vector space and therefore, the learned embeddings and their cosine similarities may not fully serve the purpose of relating the response to the input. 
\end{enumerate}
Therefore, instead of adopting the two available methods directly, we improve on the second one by fusing the latent spaces of the input and the responses, inspired by \citet{gao2019jointly}. 
Specifically, we propose to learn a linear embedding mapping from the latent space of the responses to the latent space of the input, and then select the best response by cosine similarity. 
Mathematically, we use $e_x$ to denote the input embedding and $e_y$ to denote the response embedding. 
We aim to learn a linear mapping from $e_y$ to $e_y'$, where $e_y' = (W + BI) \cdot e_y$, $W$ and $B$ are learnable parameters, and $I$ is an identity matrix. 
We learn the mapping such that the sum of the cosine similarities between $e_x$ and $e_y'$ for the training data is maximized. 
By way of this transformation, $e_y'$ now maps the vector space of the responses to that of the input, and thus allows the pretrained model to effectively utilize the discriminative power of the sentence embeddings. 
We empirically observe that the linear mapping works well and leave other advanced mapping techniques for future work.

\section{Empirical Evaluation}
\label{sec:exp}

\begin{table*}[t!]
	\centering
	\small
	\begin{tabular}{c|c|cccccc|ccccc|c}
		\toprule
		\multicolumn{2}{c}{}& \multicolumn{6}{c}{\textbf{Diversity}} &  \multicolumn{5}{|c}{\textbf{Relevance}} & \multicolumn{1}{|c}{\textbf{LQ.}} \\ 
		\cmidrule{3-14} 
		\multicolumn{2}{c}{} & \textbf{Dist-1} & \textbf{Dist-2} & \textbf{Ent-1} & \textbf{Ent-2} & \textbf{SB1*} & \textbf{SB2*}  & \textbf{B2}  & \textbf{R2} & \textbf{MS} & \textbf{BS}  & \textbf{BM25} & \textbf{GR}\\
		\midrule
		\multirow{5}{*}{\rotatebox[origin=c]{90}{\textbf{CONAN}}}
		&\textbf{Seq2Seq} & \textbf{0.06} & 0.23 & 5.12 & 6.63 & 0.54 & 0.30 & 3.4 & 3.0 &4.4 &0.83 &2.66 & 0.38\\
		&\textbf{MMI} & \textbf{0.06} & 0.23 & 4.88 & 6.41 & 0.57 & 0.35   & 2.9  & 2.3 &3.9&0.82 &1.63 & 0.33\\
		&\textbf{SpaceFusion} & 0.00 & 0.00 & 1.06 & 1.86 & 0.98 & 0.98  & 0.0  & 0.0 & -14.2 &0.76 & 0.12& 0.38\\
		&\textbf{BART} & 0.04 & 0.23 & \textbf{5.98} & \textbf{7.80} & 0.52 & 0.26 & 3.9 & 3.6 &7.1&0.84 & 1.86 &\textbf{0.71}\\
		&\textbf{\our} & \textbf{0.06} & \textbf{0.27} & 5.77 & 7.41 & \textbf{0.43} & \textbf{0.19}  & \textbf{7.1} & \textbf{6.5} & \textbf{10.9} &\textbf{0.85}  & \textbf{5.43} & \textbf{0.71} \\
		\midrule
		
		\multirow{5}{*}{\rotatebox[origin=c]{90}{\textbf{Reddit}}}
		&\textbf{Seq2Seq} & 0.04 & 0.24 & 5.07 & 6.61 & 0.58 & 0.31 & 6.5  & 4.0 &6.8&0.85& 0.14 & 0.64\\
		&\textbf{MMI} & 0.05 & 0.32 & 5.11 & 6.76 & 0.56 & 0.29  & 6.4  & 4.0 &6.9&0.85& 0.14 & 0.56\\
		&\textbf{SpaceFusion} & 0.00 & 0.02 & 2.73 & 4.16 & 0.87 & 0.76  & 0.9 & 0.0 &-2.5&0.79& 0.16 &  0.26\\
		&\textbf{BART} & 0.03 & 0.19 & 5.08 & 6.63 & 0.69 & 0.55 & 7.8 & 6.9 & \textbf{7.8} &0.86 & 0.83&  0.72\\
		&\textbf{\our} & \textbf{0.09} & \textbf{0.53} & \textbf{5.74} & \textbf{7.61} & \textbf{0.41} & \textbf{0.15} & \textbf{8.1} & \textbf{7.1} & \textbf{7.8} &\textbf{0.87}& \textbf{2.58} & \textbf{0.75}\\
		\midrule
		
		\multirow{5}{*}{\rotatebox[origin=c]{90}{\textbf{Gab}}}
		&\textbf{Seq2Seq} & 0.02 & 0.17 & 5.14 & 6.71 & 0.56 & 0.30 & 7.5 & 5.0 &6.7&0.86 & 0.14& 0.67\\
		&\textbf{MMI} & 0.02 & 0.17 & 5.28 & 6.82 & 0.55 & 0.30 & 5.8 & 3.6 &6.2&0.85& 0.18 & 0.65 \\
		&\textbf{SpaceFusion} & 0.00 & 0.01 & 3.72 & 4.84 & 0.81 & 0.73 & 1.8 & 0.1 &0.0&0.82& 0.17 &  0.21\\
		&\textbf{BART} & 0.03 & 0.17 & 5.42 & 7.25 & 0.60 & 0.38 & 6.9 & \textbf{6.4} & \textbf{6.8} &0.86 & 0.81 & 0.72\\
		&\textbf{\our} & \textbf{0.06} & \textbf{0.40} & \textbf{5.82} & \textbf{7.83} & \textbf{0.39} & \textbf{0.15} & \textbf{7.6} & \textbf{6.4} & \textbf{6.8} & \textbf{0.87} & \textbf{1.94}& \textbf{0.76} \\
		
		\bottomrule
	\end{tabular}
	\caption{Automatic evaluation results. 
		An asterisk \textbf{*} by the metric name indicates that the metric favors smaller values. 
		Best results are in bold.
		LQ.: Language Quality; SB1: Self-BLEU-1; SB2: Self-BLEU-2; B2: BLEU-2; R2: ROUGE-2; MS: MoverScore; BS: BERTScore; GR: GRUEN.}
	\label{table:results}
\end{table*}

\begin{table}[t!]
	\centering
	\small
	\begin{tabular}{c|c|ccc}
		\toprule
		\multicolumn{2}{c}{}& \textbf{Div.} & \textbf{Rel.} & \textbf{LQ.}  \\
		\midrule
		\multirow{4}{*}{\rotatebox[origin=c]{90}{\textbf{CONAN}}}
		&\textbf{Seq2Seq}  & 0.50 & 0.22    & 0.06\\
		&\textbf{MMI} & 0.55  & 0.08 & 0.02\\
		&\textbf{BART} & 0.40  & 0.73  &  0.65\\
		&\textbf{\our}  & \textbf{0.80} & \textbf{0.83} &  \textbf{0.66} \\
		\midrule
		
		\multirow{4}{*}{\rotatebox[origin=c]{90}{\textbf{Reddit}}}
		&\textbf{Seq2Seq}  & 0.25  & 0.23 & 0.38 \\
		&\textbf{MMI}  & 0.35  & 0.23  & 0.35\\
		&\textbf{BART}  & 0.00 & 0.47 & \textbf{0.51} \\
		&\textbf{\our}  & \textbf{1.00}  & \textbf{0.58} & 0.48 \\
		\midrule
		
		\multirow{4}{*}{\rotatebox[origin=c]{90}{\textbf{Gab}}}
		&\textbf{Seq2Seq} & 0.35  & 0.36   & 0.31\\
		&\textbf{MMI}  & 0.55 & 0.34 & 0.27 \\
		&\textbf{BART}  & 0.10  & 0.42 & 0.35\\
		&\textbf{\our} & \textbf{0.80}  & \textbf{0.47}  & \textbf{0.36} \\
		
		\bottomrule
	\end{tabular}
	\caption{Human evaluation results. Div.: Diversity; Rel.: Relevance; LQ.: Language Quality. }
	\label{table:human_results}
\end{table}

In this section, we empirically evaluate the performance of our proposed approach and a set of baseline models. 

\subsection{Experimental Setup}
\label{sec:setup}
\noindent \textbf{Datasets}: 
We use the benchmark datasets collected by \citet{qian2019benchmark}, which are fully-labeled hate speech intervention datasets collected from Reddit and Gab, comprising 5,257 and 14,614 hate speech instances respectively. 
We use the filtered conversation setting in \citet{qian2019benchmark}, which includes the posts labeled as hate speech only and discards other non-hateful conversations. 
Besides, we use the English language portion of the CONAN dataset \cite{chung2019conan}, which contains counterspeech for 408 hate speech instances, written by experts trained on countering hatred. 
The Reddit, Gab and CONAN datasets have on average 2.66, 2.86 and 9.47 ground truth counterspeech for each hate speech respectively.

\noindent \textbf{Training Data}: 
Since each hate speech can have multiple ground truth counterspeech, we follow \citet{qian2019benchmark} to dis-aggregate the counterspeech and construct a pair (hate speech, counterspeech) for each of the ground truth counterspeech in each dataset. 
Given a counterspeech dataset, we randomly choose 70\% (hate speech, counterspeech) pairs for model training, 15\% for cross validation and the rest 15\% for testing.

\noindent \textbf{Baselines}: We compare our proposed approach with the following competitive baseline models:
\begin{enumerate}[leftmargin=*]
	\setlength\itemsep{-0.2em}
	\item Seq2Seq \cite{sutskever2014sequence,cho2014learning} is a widely used neural model for language generation. We use 2 bidirectional Gated Recurrent Unit (GRU) layers for the encoder and 2 GRU layers followed by a 3-layer neural network as the decoder. 
	\item Maximum Mutual Information (MMI) \cite{li2016diversity} is a diversity-promoting approach for neural conversation models. 
	We implement the MMI-bidi model \cite{li2016diversity} and adopt incremental learning \cite{ranzato2016sequence} to facilitate robust training. 
	\item SpaceFusion \cite{gao2019jointly} 
	optimizes both diversity and relevance by introducing a fused latent space, where the direction and distance from the predicted response vector roughly match the relevance and diversity, respectively. 
	We align the direction parameter with the ground truth counterspeech. 
	To better exercise the diversity power, we randomly choose the distance parameter at each time of generation. 
	\item BART \cite{lewis2020bart} 
	is the state-of-the-art pre-trained sequence-to-sequence model for language generation. 
	It has a standard Transformers-based neural machine translation architecture which can be seen as generalizing BERT \cite{devlin2019bert}, GPT \cite{radford2018improving} and many pretraining schemes. 
	We fine-tune the BART model on our training data. 
\end{enumerate}

\noindent We compare with Seq2Seq since they are initially proposed and used by \citet{qian2019benchmark}.\footnote{We do not include the results of the variational auto-encoder model and the reinforcement learning model in \citet{qian2019benchmark} for comparison as they has very similar performance as Seq2Seq. Readers are referred to \citet{qian2019benchmark} for detailed performance.} 
We select MMI, SpaceFusion and BART as baselines because they are the state-of-the-art models in promoting diversity, optimizing both diversity and relevance, and generating quality language respectively.

\subsection{Evaluation}
We evaluate all model outputs along three dimensions: diversity, relevance and language quality. 
\textit{Diversity} refers to vocabulary richness, variety in expression and the extent to which the response is dissimilar from the rest  in a generated collection of responses. 
\textit{Relevance} captures the extent to which the counterspeech addresses the central aspect of the hateful message and makes a coherent conversation towards mitigating the hate speech.
A low relevance score means that the counterspeech is irrelevant to the hate speech or off-topic (\eg, the hate speech talks about LGBTQ whereas the counterspeech is related to religious beliefs). 
\textit{Language quality} measures whether the generated responses are grammatical, fluent and readable.

\subsubsection{Automatic Evaluation}
We evaluate \textit{diversity} by distinct n-grams (\textbf{Dist-n}) \cite{li2016diversity}, Entropy (\textbf{Ent-n}) \cite{zhang2018generating} and \textbf{Self-BLEU} \cite{zhu2018texygen}. 
For \textit{relevance}, we compare 1) the generated response with the ground truth counterspeech by \textbf{BLEU} \cite{papineni2002bleu} and \textbf{ROUGE} \cite{lin2003automatic,lin2004rouge} for syntactic similarity, and by \textbf{MoverScore} \cite{zhao2019moverscore} and \textbf{BERTScore} \cite{zhang2020bertscore} for semantic similarity; 
2) the generated response with the hate speech by \textbf{BM25} \cite{manning2008introduction}, a relevance estimation function widely used in information retrieval. 
We adopt GRUEN \cite{zhu2020gruen} to evaluate the \textit{language quality}. 
Note that larger scores indicate better quality, except for Self-BLEU. 

\subsubsection{Human Evaluation}
Following \citet{qian2019benchmark}, we ask human annotators to perform ``winner selection'' instead of numeric ratings. 
We randomly sampled 100 hate speech instances from each dataset's test collection, and corresponding model-generated counterspeech outputs from Seq2Seq, MMI, BART, and our model.\footnote{To save human annotation effort, we asked human raters to annotate only the best 4 models w.r.t. automatic metrics, and therefore leave SpaceFusion out of annotation.} 
Two human annotators (native English speakers and sensitive about hate speech online) were presented with one hate speech and the four generated outputs, and were asked to evaluate the outputs for relevance and language quality separately. 
As for diversity, the annotators were presented with 10 responses for each of the four models each time, and were asked to select the model with the best diversity. 
For each of the three dimensions, the \textit{best} output received a $1$ (ties were allowed, \ie, multiple outputs could receive a $1$) and the rest received a $0$.

\subsection{Results}

\begin{table*}
	\small
	\centering
	\begin{tabular}{*{3}{c}}
		\toprule
		\textbf{Hate Speech} & \textbf{Generated Counterspeech by \our}\\
		\midrule
		\multicolumn{1}{p{0.47\textwidth}}{\textbf{(a)}: All the Muslims should have been punished for yesterday, not only a few. Citizen will take action if the government does not do anything.}& 
		\multicolumn{1}{p{0.48\textwidth}}{You cannot blame all people for actions of a few. Banning something altogether will not solve anything.} \\
		\midrule
		\multicolumn{1}{p{0.47\textwidth}}{\textbf{(b)}: The world would be a better place without Muslims. They are only killing and raping our children.}& 
		\multicolumn{1}{p{0.48\textwidth}}{The world would actually be a very dark place without Muslims who contribute a lot to our society. What about our Muslim doctors, scientists, philanthropists, actors, job-creators?}\\
		\midrule
		\multicolumn{1}{p{0.47\textwidth}}{\textbf{(c)}: What is the point in listening to black women? They are all retarded.}& 
		\multicolumn{1}{p{0.48\textwidth}}{The point of listening to black women, and women in general, is that they could broaden and enlighten your view of the world.}\\
		\bottomrule
	\end{tabular}
	\caption{Examples of the generated counterspeech}
	\label{table:case_study}
\end{table*}

We obtain the results of our model (denoted as ``GPS") by generating 30k, 30k, 40k candidates for the CONAN, Reddit and Gab datasets respectively.\footnote{The number of generated candidates are selected based on the performance on the cross validation datasets.} 
After the Candidate Pruning module, we have 15.4k, 17.9k and 25.4k grammatical candidates for each dataset respectively. 

The results by automatic metrics and human evaluation metrics are presented in Table \ref{table:results} and Table \ref{table:human_results} respectively. 
Overall, \our has the best diversity with significant margins than the baselines.
For relevance, \our has slightly better performance for BLEU, ROUGE, MoverScore and BERTScore, while has much better performance on BM25. 
This implies the counterspeech generated by \our are more related to the hate speech and therefore, make more coherent conversations. 
Examples of counterspeech generated by \our are presented in Table \ref{table:case_study}. 
We find that \our is able to generate diverse and relevant rather than merely commonplace responses, such as ``Please refrain from using such language". 
Comparative case studies for different baseline models are shown in Appendix \ref{sec:app_case_study}. 
Therefore, we conclude that \our has the best diversity and relevance, compared to the baselines. 
Besides, \our has comparable language quality with the best baseline model---BART. 


Among these baselines, BART is the strongest one with much better relevance and language quality. 
Yet, BART still suffers from the diversity issue, as discussed in Section \ref{sec:Conversational_Language_Generation}. 
SpaceFusion has very poor results overall, though a manual inspection of the latent space fusion visualization suggests otherwise.
One explanation is that SpaceFusion, with substantially more parameters compared with the Seq2Seq model may not have had sufficient training instances for its optimal performance. 
In their own experiments, \citet{gao2019jointly}, demonstrate that SpaceFusion worked well on two datasets with 0.2M and 7.3M conversations, which is at least one to two orders of magnitude larger than our dataset.
If provided with more training data, SpaceFusion could possibly be a strong candidate too. 
In comparison, though BART is an even more complicated model with 139M parameters, it was pre-trained on the BooksCorpus dataset \cite{zhu2015aligning} with over 7,000 unique unpublished books and has the fine-tunable property. 

\subsection{Ablation Study}
\label{sec:ablation}
\begin{figure}[ht]
	\centering
	\includegraphics[width=0.99\linewidth]{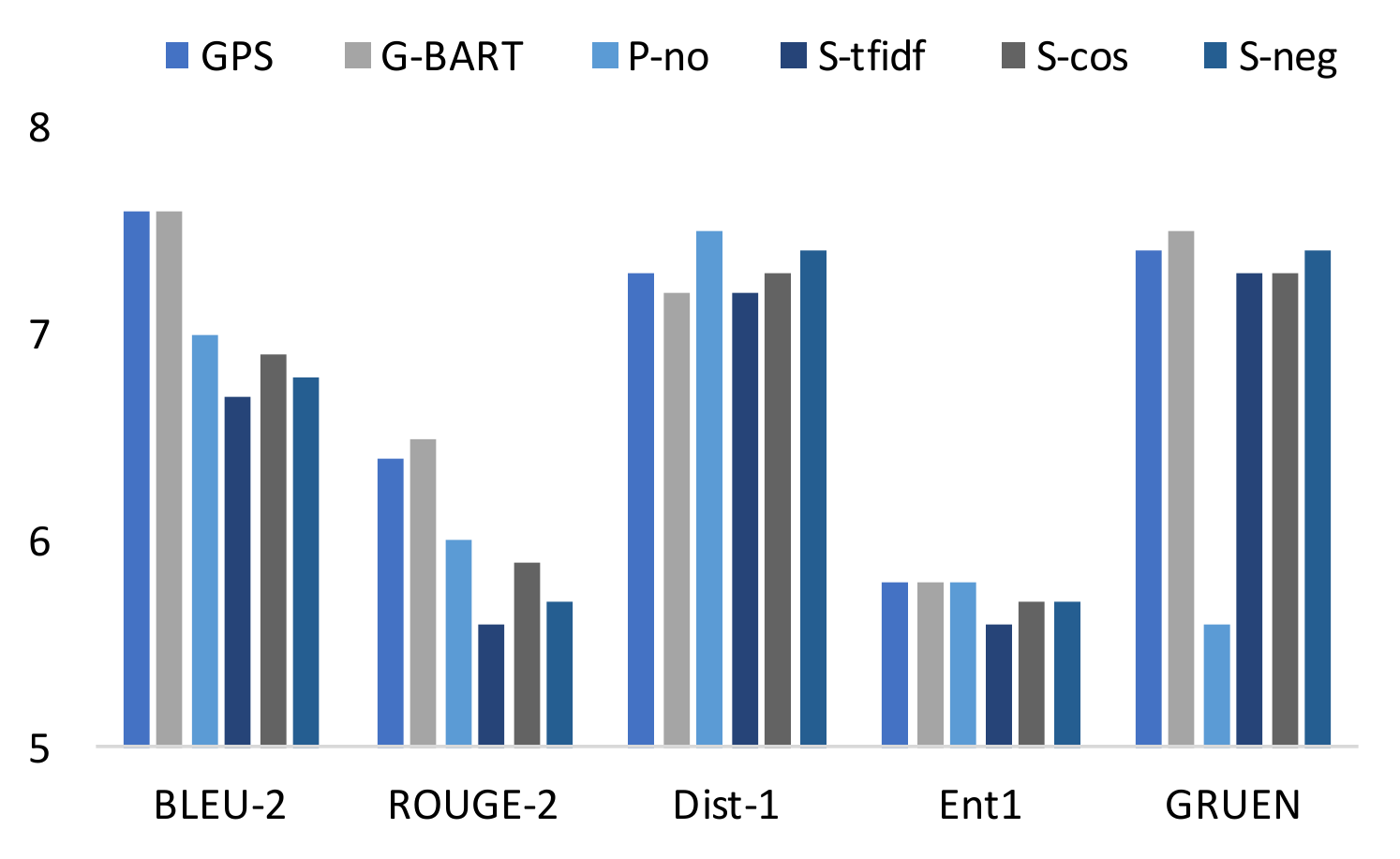}
	\caption{Ablation study. Plots show average results across all three datasets. 
		We scale Dist-1 by 100 times and GRUEN by 10 times for better visualization.}
	\label{fig:ablation}
\end{figure}

We compare with the following ablations of \our and show the results in Figure \ref{fig:ablation}. 
\begin{enumerate}[leftmargin=*]
	\setlength\itemsep{-0.2em}
	\item G-BART: instead of generating the candidates by the RNN-based variational autoencoder  \cite{bowman2016generating}, we generate the candidates by BART \cite{lewis2020bart}. 
	\item P-no: we exclude the pruning module and make all generated candidates available for selection. 
	\item S-tfidf: we select the most relevant response by tf-idf on raw texts. 
	\item S-cos: we exclude the latent space fusion step and select the best response by the cosine similarity of the response embeddings and the hate speech embeddings \cite{henderson2019training}.
	\item S-neg: we use the negative sampling technique to train a response selection classifier \cite{henderson2019training}. 
\end{enumerate}

G-BART has almost the same performance as \our. Therefore, we select the RNN-based variational autoencoder for candidate generation for its better computational efficiency. 
Compared with the full model, though P-no has slightly better performance on diversity, it performs poorly on both relevance and language quality. 
Three ablation methods for response selection have similar performance. 
They have comparable performance to \our on diversity and language quality, but worse results on relevance. 

The ablation study demonstrates the significance of the Candidate Pruning module and our proposed Response Selection method. 
It also implies that diversity, language quality and relevance are improved by the Candidate Generation module, the Candidate Pruning module, and the Response Selection module respectively.

\subsection{Generation \textit{vs.} Selection}
\label{sec:generated_vs_selection}
\begin{figure}[ht]
	\centering
	\includegraphics[width=0.6\linewidth]{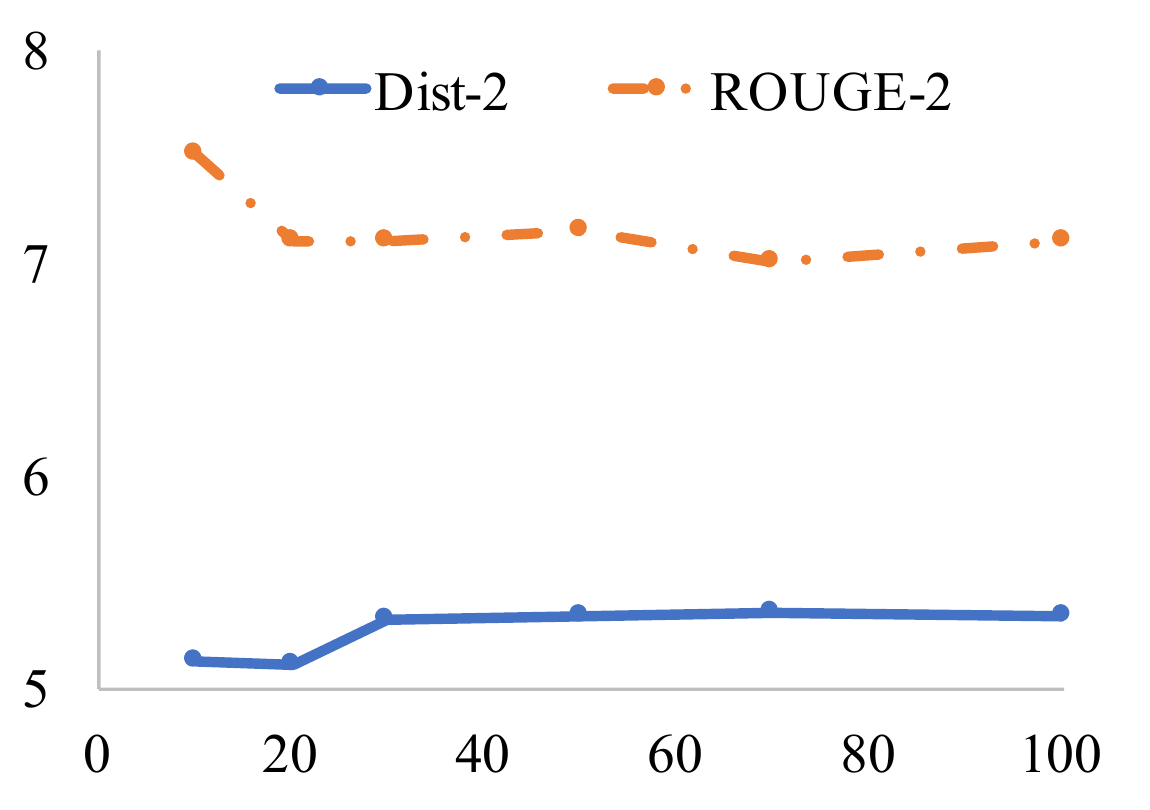}
	\caption{Dist-2 and ROUGE-2 \textit{vs.} Number of candidates (in thousands) generated on the Reddit dataset. 
		We scale Dist-2 by 10 times for better visualization.
	}
	\label{fig:ablation3}
\end{figure}

This section studies the relationship between the Candidate Generation module and the Response Selection module. 
The more candidates we generate, the more diversity the model gains potentially. 
However, one might think that the selection model may suffer from a very large candidate pool and result in poor relevance. 
Empirically as shown in Figure \ref{fig:ablation3}, we find that once the number of candidates generated has passed a threshold, the diversity (\ie, the blue line) almost converges. 
Besides, we also find the relevance is not compromised and relatively stable even with more candidates generated beyond the threshold. 
Therefore, we select the number of candidates at the ``elbow'' point based on the performance on the validation dataset, for efficient computations. 

\subsection{Explicit Relevance (BM25) \textit{vs.} Diversity}
\begin{figure}[ht]
	\centering
	\includegraphics[width=0.49\linewidth]{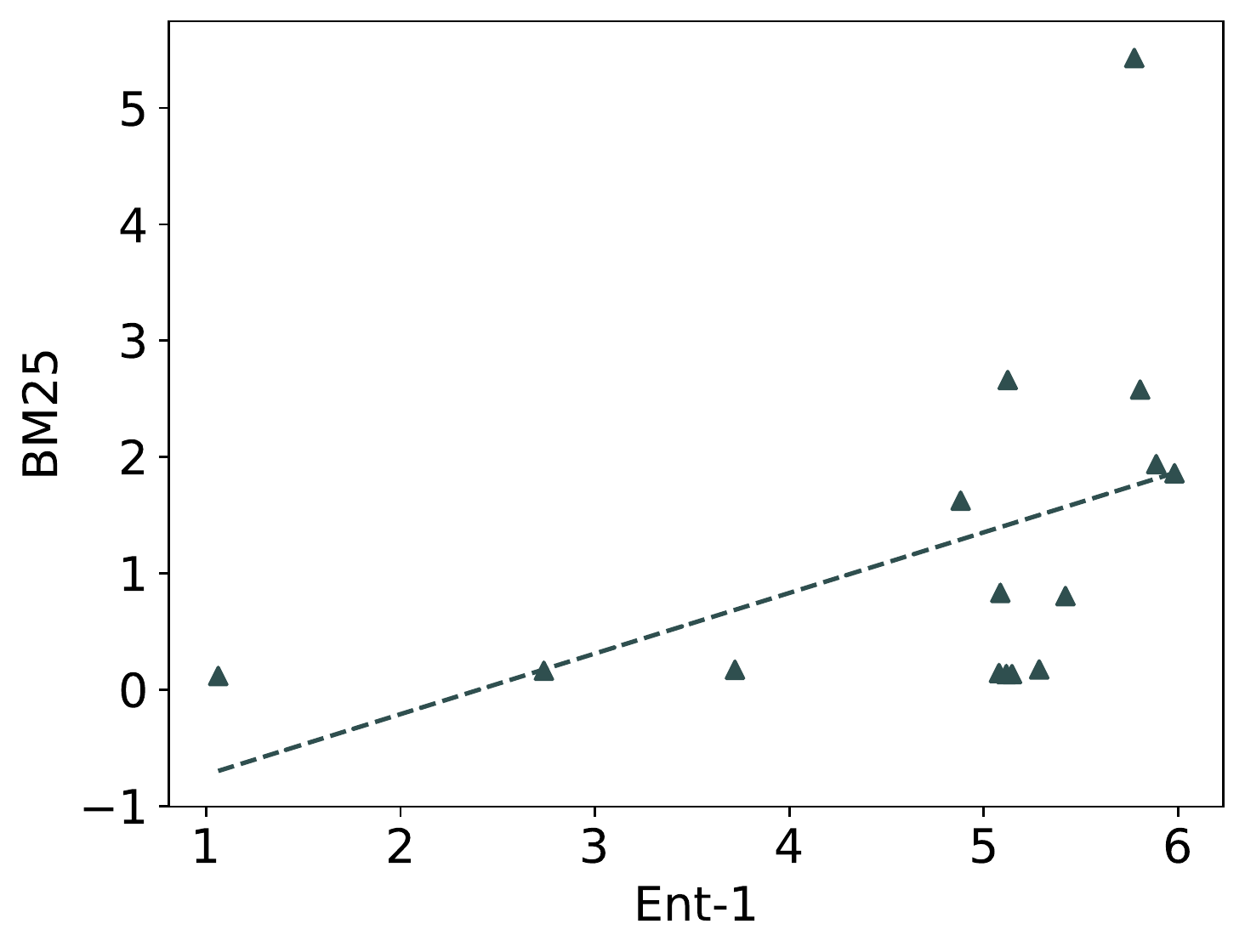}
	\includegraphics[width=0.49\linewidth]{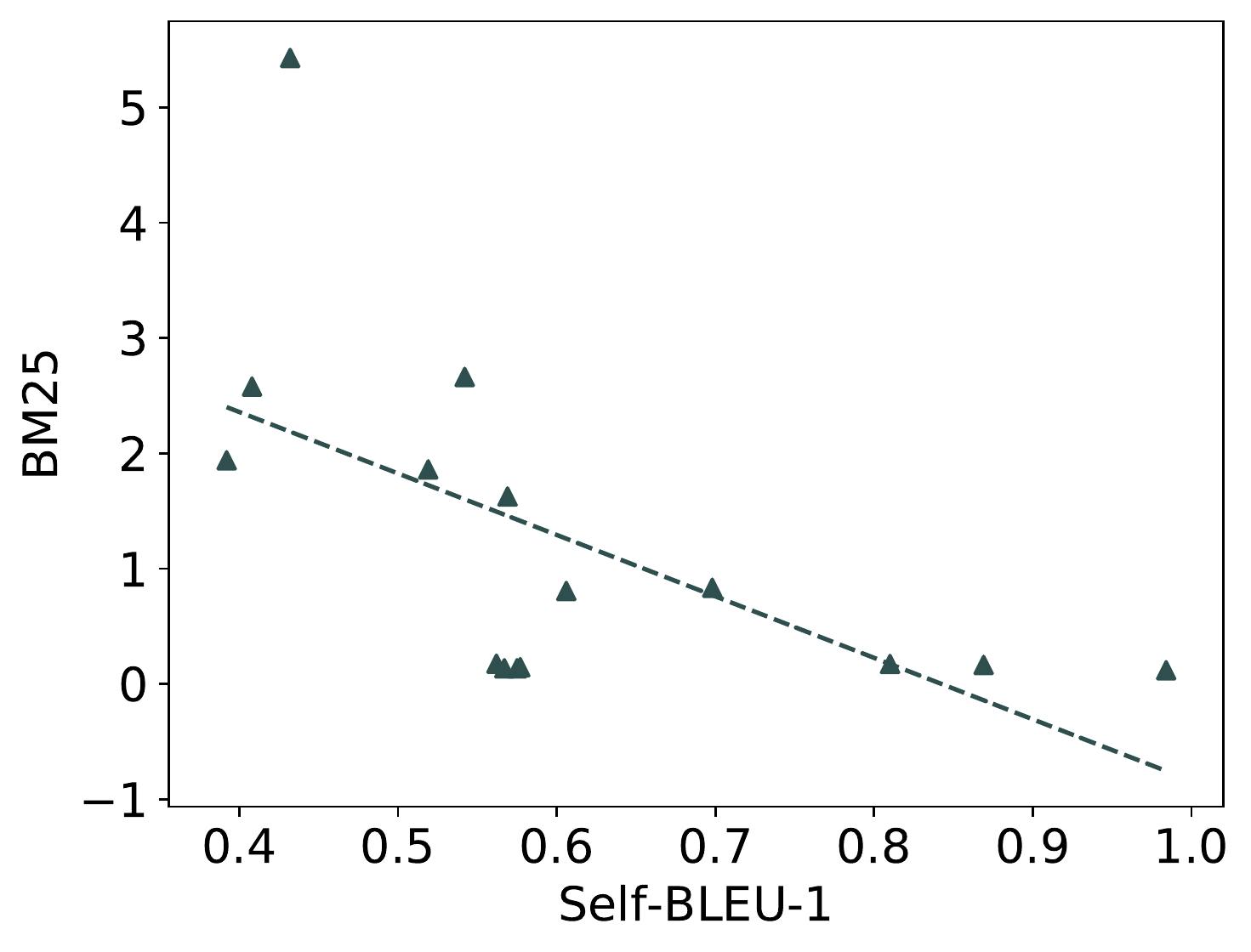}
	\caption{BM25 \textit{vs.} Diversity. 
		Each data point denotes a (diversity, BM25) pair for one model on one dataset, and the dotted lines indicate regression lines. 
	}
	\label{fig:relevance_vs_diversity}
\end{figure}

Based on the reasoning that models with better BM25 scores should specifically address the central aspect of hate speech and thus produce dissimilar responses for different hate speech, we hypothesize that models with better BM25 should generate more diverse responses. 
Therefore, we present scatter plots of BM25 and diversity scores for all five models (in Section \ref{sec:setup}) on all three datasets altogether in Figure \ref{fig:relevance_vs_diversity}, resulting 15 data points per subfigure. 
We find that BM25 and diversity have a reasonably strong correlation (Pearson's Correlation scores are 0.47 and -0.60 for Ent-1 and Self-BLEU-1 respectively). 

\section{Related Work}
\label{sec:related_work}
We focus on three areas to the problem of hate speech and its countermeasures, \ie (i) psychosocial analysis, (ii) automatic counterspeech generation, and more broadly, (iii) conversational language generation.

\subsection{Psychosocial Analysis of Counterspeech}
\label{sec:psychoanalysis}
\textbf{Effectiveness of Counterspeech}: 
There is a significant research interest in understanding the effectiveness of counterspeech to fight hatred and de-escalate the conversation as evidenced by a growing number of recent studies \cite{schieb2016governing,munger2017tweetment,mathew2018analyzing}. 
\citet{munger2017tweetment} found that subjects who were educated by high-follower white males, significantly reduced their use of racist slurs on Twitter. 
\citet{schieb2016governing} studied counterspeech on Facebook via a simulation, and concluded that counterspeech could have a considerable impact on a given audience, and the impact was a function of the proportion of hate speakers in the audience. 
In a subsequent study, \citet{mathew2018analyzing} recorded the case of a user who, after seeing the counterspeech  posted to her  hateful messages on Twitter, openly apologized for her actions.
Besides academia, some organizations are also set to promote countermeasures via campaigns such as the no hate speech movement\footnote{\url{https://www.nohatespeechmovement.org}} and the Facebook counterspeech campaign\footnote{\url{https://counterspeech.fb.com}}. 
Therefore, \citet{benesch2014countering,mathew2019thou} suggest that counterspeech can be regarded as one of the most promising and ``constitutionally preferred" approaches to hate speech. 
In addition, counterspeech could be likened to the effect of prosocial active bystanders in face-to-face bullying scenarios, where bystander intervention (speaking on behalf of the victim) has been found to successfully abate victimization most of the time \cite{o1999peer,craig2000observations}.

\noindent \textbf{Psychosocial and Linguistic Aspects}:
Besides the effectiveness of counterspeech, psychosocial and linguistic aspects of both counterspeech and hate speech have been actively studied by \citet{mathew2019spread,siegel2018online,schieb2016governing,weingartner2019online,mathew2018analyzing}.
For instance, \citet{mathew2019thou} performed detailed psycholinguistic analysis on counterspeech, compared the effectiveness of different counterspeech strategies, and revealed some important insights, such as counterspeech comments receive much more ``likes" on YouTube compared to the non-counterspeech comments, suggesting a communal empathy for the target of hate speech. 
Besides, \citet{mathew2019thou,chung2019conan} studied different strategies (\eg, call for influential users) to produce effective counterspeech. 
\citet{mathew2018analyzing} found that the hate tweets by verified accounts were much more viral as compared to  tweets by non-verified accounts, by analyzing the hate speech and counterspeech accounts on Twitter. 
\citet{mathew2019spread} study how hate speech spreads in online social media.  
More recently, \citet{sap2020social} studied pragmatic formalisms to capture ways in which people express social biases and power differentials in language, permitting a broader computational framework for processing hate speech.

\subsection{Counterspeech Generation}
Though the effectiveness of counterspeech is well-motivated from both psychosocial and linguistic perspectives, limits to manual counterspeech generation at scale have prompted automatic generation of counterspeech, an area that has received little attention to date. 
The first key challenge in this direction is the creation of reliable counterspeech datasets of high quality. 
\citet{mathew2019thou} collected counterspeech from YouTube comments, but omit the hate speech associated with each counterspeech. 
Such a dataset may be good for psychosocial and linguistic analysis, but is not sufficient for training an NLG model. 
To enable model training, \citet{qian2019benchmark} released two fully-labeled datasets collected from Reddit and Gab. 
Besides, \citet{chung2019conan} collected a quality dataset where the counterspeech instances are written by trained experts and are meant to fight each hate speech and de-escalate a hateful situation. 
Recently, \citet{tekiroglu2020generating} proposed an approach to collect counterspeech responses in a more effective manner, but have not yet released a quality dataset. 
In our work, we conduct the experiments on all the publicly available datasets (\ie, \cite{chung2019conan,qian2019benchmark}) to date, to the best of our knowledge.

Research on NLG algorithms for counterspeech generation is still in its infancy. \citet{qian2019benchmark} made the only initial attempt and proposed the use of three neural models to generate counterspeech. 
However, they only experimented with the most basic model architectures (\eg, Seq2Seq) to prove the feasibility of the task, and leave the performance improvement for future work. 
In our work, we extend their results by 
studying more advanced architectures, identifying principal dimensions of effective counterspeech, and proposing a novel pipeline to better solve the problem. 
To the best of our knowledge, this paper represents the first successful pipeline model for counterspeech generation.

From the technical perspective, our work shares some high-level similarities with \citet{tekiroglu2020generating} since we both use generative models to generated candidates. 
However, we would like to highlight that our essential goals are different. 
\citet{tekiroglu2020generating} aim to \textit{collect quality data} by enabling language models and studying human  annotation strategies, while we aim to \textit{generate counterspeech} to a given hate speech.

\subsection{Conversational Language Generation}
\label{sec:Conversational_Language_Generation}
Counterspeech generation is broadly related to conversational language generation, where most of the best performing approaches are based on neural models trained in a sequence-to-sequence manner \cite{see2019massively}. 
Despite the good performance of these models, one of their widely acknowledged intrinsic drawbacks is the generation of safe and commonplace responses 
\cite{sordoni2015neural} 
due to improper objective function \cite{li2016diversity}, 
lack of model variability \cite{serban2017hierarchical,zhao2017learning}, 
weak conditional signal \cite{tao2018get},
and model over-confidence \cite{jiang2018sequence}. 
Such tendency has prompted the study of methods that improve diversity and has resulted in a wide variety of solutions, such as 
optimizing a different loss function \cite{li2016diversity,zhang2018generating}, 
varying the latent space \cite{shao2019long,gao2019jointly}, 
utilizing adversarial learning \cite{xu2018diversity,shetty2017speaking,shi2018toward},
and leveraging non-conversational information \cite{wu2020diverse,su2020diversifying,tu2019generating}. 
Our work is different from all above in that we adopt a \textit{pipeline} model which promotes diversity by generating a variety of candidates. 
As such, it does not have the aforementioned intrinsic drawback of a sequence-to-sequence model.

\section{Conclusion and Future Work}
\label{sec:conclusion}
We proposed a three-module pipeline --- \textit{Generate}, \textit{Prune}, \textit{Select} for counterspeech generation against online hate speech. 
Empirical evaluation on three datasets demonstrates that our model is effective in producing diverse and relevant counterspeech. 

Future works could include the following two directions: 
1) stylistic counterspeech generation: 
\citet{mathew2019thou} find that different counterspeech styles/strategies may be needed for different hate speech topics and therefore, it would be interesting to develop new techniques to generate the most effective style of counterspeech for each hate topic. 
We think this could be a natural extension to our proposed model, since we can utilize a style classifier in the Candidate Pruning module. 
2) system deployment: studying the real social impacts of automatic counterspeech generation in reducing online hate speech via system deployment and the actual activity monitoring can directly inform research in this area.


\noindent \textbf{Reproducibility}: Our code is available at \url{https://github.com/WanzhengZhu/GPS}.

\section*{Acknowledgments}
We thank the anonymous reviewers for comments on earlier drafts that significantly helped improve this manuscript. 
We thank Pooja Bhagchandani and Neha Prabhu for annotating the model outputs. 
This research was supported by the National Science Foundation award CNS-1720268.

\section*{Ethical Considerations}

We recognize that studying counterspeech generation necessarily requires us to confront online content that may be offensive or disturbing. 
However, deliberate avoidance does not eliminate such problems \cite{sap2020social}. 
Since the effectiveness of counterspeech has already been widely studied in Section~\ref{sec:psychoanalysis}, our work makes a positive step towards automating the process, which could potentially educate hate speakers and mitigate hate speech online. 
Besides, the automation process could help reduce the amount of human work and therefore, potential harm to human moderators \cite{barrett2020moderates,zhu2021selfsupervised}. 
In addition, the collective analysis over large corpora and counterspeech can also be insightful for educating people on reducing the usage of hate speech consciously or unconsciously in their language.

\noindent \textbf{Risks in deployment}: 
The deployment of counterspeech generation (\eg, \cite{de2020toxicbot}) should be done after paying attention to several  ethical aspects  some of which we list below. 
\begin{itemize}[leftmargin=*]
	\setlength\itemsep{-0.2em}
	\item Social and racial bias \cite{sap2020social}: Does the model have any pragmatic implications which project unwanted social or racial biases and stereotypes onto online users? 
	\item Fairness \cite{mitchell2019model,corbett2017algorithmic}: can the model ensure fairness for different demographic groups or speakers of different forms/dialects/vernaculars of English? 
	\item Failure cases: are there any failure cases, which could further incite more aggressive hate speech? It is crucial to ensure that  counterspeech deployment does not escalate a given hateful situation. 
	\item Evaluation metrics \cite{corbett2017algorithmic}: the present study improves upon prior works by more comprehensive evaluations on diversity, relevance and language quality. However, there is a chance that the three criteria are sufficient for deployment in a realistic setting and  there may be additional criteria associated with their effectiveness. 
	\item Potential nefarious side effects and misuse potential \cite{lau2020give}: how to ensure that our model is not misused for other unwanted purposes? 
\end{itemize}
Given the limited scope of the present study, we call for attention to these aspects by way of well-designed experiments before deploying counterspeech generation bots.

\noindent \textbf{Regulatory standpoint on the present study}: 
Institutional Review Board (IRB) gave us clear feedback on what is considered human research and thus subject to IRB review. 
Analyses relying on user-generated content do not constitute human-subject research, and are thus not the purview of the IRB, as long as 
1) the data analyzed are posted on public fora and were not the result of direct interaction from the researchers with the people posting, 
2) there are no private identifiers or personally identifiable information associated with the data, 
and 3) the research is not correlating different public sources of data to infer private data.\footnote{This position is in line with Title 45 of the Code of Federal Regulations, Part 46 (45 CFR 46), which defines human research.} 
All of these conditions apply to the present study. Additionally, the hate speech and counterspeech instances were secondary data, previously collected by \citet{qian2019benchmark,chung2019conan} and the annotators in our study were evaluating the quality of the generated sentences only.

\noindent \textbf{Risks in annotation}: 
The data we use in this paper were posted on publicly accessible websites, and do not contain any personally identifiable information (i.e., no real names, email addresses, IP addresses, etc.). 
The annotators were undergraduate assistants in the lab receiving research credit for their annotation and were blind to the systems they were annotating. They were warned about the offensive content before they read the data, and were informed that they could quit the task at any time if they were uncomfortable with the content.

\bibliographystyle{acl_natbib}
\bibliography{ref}

\begin{thebibliography}{65}
\expandafter\ifx\csname natexlab\endcsname\relax\def\natexlab#1{#1}\fi

\bibitem[{Barrett(2020)}]{barrett2020moderates}
Paul~M. Barrett. 2020.
\newblock Who moderates the social media giants?
\newblock \emph{Center for Business}.

\bibitem[{Benesch(2014)}]{benesch2014countering}
Susan Benesch. 2014.
\newblock Countering dangerous speech: New ideas for genocide prevention.
\newblock \emph{Washington, DC: United States Holocaust Memorial Museum}.

\bibitem[{Bowman et~al.(2016)Bowman, Vilnis, Vinyals, Dai, Jozefowicz, and
  Bengio}]{bowman2016generating}
Samuel Bowman, Luke Vilnis, Oriol Vinyals, Andrew Dai, Rafal Jozefowicz, and
  Samy Bengio. 2016.
\newblock Generating sentences from a continuous space.
\newblock In \emph{Proceedings of the 20th SIGNLL Conference on Computational
  Natural Language Learning}.

\bibitem[{Cho et~al.(2014)Cho, van Merri{\"e}nboer, Gulcehre, Bahdanau,
  Bougares, Schwenk, and Bengio}]{cho2014learning}
Kyunghyun Cho, Bart van Merri{\"e}nboer, Caglar Gulcehre, Dzmitry Bahdanau,
  Fethi Bougares, Holger Schwenk, and Yoshua Bengio. 2014.
\newblock Learning phrase representations using rnn encoder--decoder for
  statistical machine translation.
\newblock In \emph{Proceedings of the Empirical Methods in Natural Language
  Processing (EMNLP)}.

\bibitem[{Chung et~al.(2019)Chung, Kuzmenko, Tekiroglu, and
  Guerini}]{chung2019conan}
Yi-Ling Chung, Elizaveta Kuzmenko, Serra~Sinem Tekiroglu, and Marco Guerini.
  2019.
\newblock Conan-counter narratives through nichesourcing: a multilingual
  dataset of responses to fight online hate speech.
\newblock In \emph{Proceedings of ACL}.

\bibitem[{Corbett-Davies et~al.(2017)Corbett-Davies, Pierson, Feller, Goel, and
  Huq}]{corbett2017algorithmic}
Sam Corbett-Davies, Emma Pierson, Avi Feller, Sharad Goel, and Aziz Huq. 2017.
\newblock Algorithmic decision making and the cost of fairness.
\newblock In \emph{Proceedings of the 23rd ACM SIGKDD International Conference
  on Knowledge Discovery and Data Mining (KDD)}.

\bibitem[{Craig et~al.(2000)Craig, Pepler, and Atlas}]{craig2000observations}
Wendy~M Craig, Debra Pepler, and Rona Atlas. 2000.
\newblock Observations of bullying in the playground and in the classroom.
\newblock \emph{School Psychology International}, 21(1):22--36.

\bibitem[{Devlin et~al.(2019)Devlin, Chang, Lee, and
  Toutanova}]{devlin2019bert}
Jacob Devlin, Ming-Wei Chang, Kenton Lee, and Kristina Toutanova. 2019.
\newblock Bert: Pre-training of deep bidirectional transformers for language
  understanding.
\newblock In \emph{Proceedings of NAACL-HLT}.

\bibitem[{Gagliardone et~al.(2015)Gagliardone, Gal, Alves, and
  Martinez}]{gagliardone2015countering}
Iginio Gagliardone, Danit Gal, Thiago Alves, and Gabriela Martinez. 2015.
\newblock \emph{Countering online hate speech}.
\newblock Unesco Publishing.

\bibitem[{Gao et~al.(2019)Gao, Lee, Zhang, Brockett, Galley, Gao, and
  Dolan}]{gao2019jointly}
Xiang Gao, Sungjin Lee, Yizhe Zhang, Chris Brockett, Michel Galley, Jianfeng
  Gao, and Bill Dolan. 2019.
\newblock Jointly optimizing diversity and relevance in neural response
  generation.
\newblock In \emph{Proceedings of NAACL-HLT}.

\bibitem[{Henderson et~al.(2019)Henderson, Vuli{\'c}, Gerz, Casanueva,
  Budzianowski, Coope, Spithourakis, Wen, Mrk{\v{s}}i{\'c}, and
  Su}]{henderson2019training}
Matthew Henderson, Ivan Vuli{\'c}, Daniela Gerz, I{\~n}igo Casanueva, Pawe{\l}
  Budzianowski, Sam Coope, Georgios Spithourakis, Tsung-Hsien Wen, Nikola
  Mrk{\v{s}}i{\'c}, and Pei-Hao Su. 2019.
\newblock Training neural response selection for task-oriented dialogue
  systems.
\newblock In \emph{Proceedings of the Association for Computational Linguistics
  (ACL)}.

\bibitem[{Jiang and de~Rijke(2018)}]{jiang2018sequence}
Shaojie Jiang and Maarten de~Rijke. 2018.
\newblock Why are sequence-to-sequence models so dull?
\newblock \emph{Proceedings of the Empirical Methods in Natural Language
  Processing (EMNLP)}.

\bibitem[{Kullback and Leibler(1951)}]{kullback1951information}
Solomon Kullback and Richard~A Leibler. 1951.
\newblock On information and sufficiency.
\newblock \emph{The Annals of Mathematical Statistics}, 22(1):79--86.

\bibitem[{Lau et~al.(2020)Lau, Baldwin et~al.}]{lau2020give}
Jey~Han Lau, Timothy Baldwin, et~al. 2020.
\newblock Give me convenience and give her death: Who should decide what uses
  of nlp are appropriate, and on what basis?
\newblock In \emph{Proceedings of the 58th Annual Meeting of the Association
  for Computational Linguistics}.

\bibitem[{Lewis et~al.(2020)Lewis, Liu, Goyal, Ghazvininejad, Mohamed, Levy,
  Stoyanov, and Zettlemoyer}]{lewis2020bart}
Mike Lewis, Yinhan Liu, Naman Goyal, Marjan Ghazvininejad, Abdelrahman Mohamed,
  Omer Levy, Ves Stoyanov, and Luke Zettlemoyer. 2020.
\newblock Bart: Denoising sequence-to-sequence pre-training for natural
  language generation, translation, and comprehension.
\newblock In \emph{Proceedings of the Association for Computational Linguistics
  (ACL))}.

\bibitem[{Li et~al.(2016)Li, Galley, Brockett, Gao, and
  Dolan}]{li2016diversity}
Jiwei Li, Michel Galley, Chris Brockett, Jianfeng Gao, and Bill Dolan. 2016.
\newblock A diversity-promoting objective function for neural conversation
  models.
\newblock In \emph{Proceedings of NAACL-HLT}.

\bibitem[{Li et~al.(2020)Li, Li, Bi, Liu, and Lam}]{li2020relevance}
Xin Li, Piji Li, Wei Bi, Xiaojiang Liu, and Wai Lam. 2020.
\newblock Relevance-promoting language model for short-text conversation.
\newblock In \emph{Proceedings of the AAAI Conference on Artificial
  Intelligence (AAAI)}.

\bibitem[{Lin(2004)}]{lin2004rouge}
Chin-Yew Lin. 2004.
\newblock Rouge: A package for automatic evaluation of summaries.
\newblock In \emph{Text Summarization Branches Out}.

\bibitem[{Lin and Hovy(2003)}]{lin2003automatic}
Chin-Yew Lin and Eduard Hovy. 2003.
\newblock Automatic evaluation of summaries using n-gram co-occurrence
  statistics.
\newblock In \emph{North American Association for Computational Linguistics
  (NAACL)}.

\bibitem[{Manning et~al.(2008)Manning, Sch{\"u}tze, and
  Raghavan}]{manning2008introduction}
Christopher~D Manning, Hinrich Sch{\"u}tze, and Prabhakar Raghavan. 2008.
\newblock \emph{Introduction to information retrieval}.
\newblock Cambridge University Press.

\bibitem[{Mathew et~al.(2019{\natexlab{a}})Mathew, Dutt, Goyal, and
  Mukherjee}]{mathew2019spread}
Binny Mathew, Ritam Dutt, Pawan Goyal, and Animesh Mukherjee.
  2019{\natexlab{a}}.
\newblock Spread of hate speech in online social media.
\newblock In \emph{Proceedings of the 10th ACM Conference on Web Science}.

\bibitem[{Mathew et~al.(2018)Mathew, Kumar, Goyal, Mukherjee
  et~al.}]{mathew2018analyzing}
Binny Mathew, Navish Kumar, Pawan Goyal, Animesh Mukherjee, et~al. 2018.
\newblock Analyzing the hate and counter speech accounts on twitter.
\newblock \emph{arXiv preprint arXiv:1812.02712}.

\bibitem[{Mathew et~al.(2019{\natexlab{b}})Mathew, Saha, Tharad, Rajgaria,
  Singhania, Maity, Goyal, and Mukherjee}]{mathew2019thou}
Binny Mathew, Punyajoy Saha, Hardik Tharad, Subham Rajgaria, Prajwal Singhania,
  Suman~Kalyan Maity, Pawan Goyal, and Animesh Mukherjee. 2019{\natexlab{b}}.
\newblock Thou shalt not hate: Countering online hate speech.
\newblock In \emph{Proceedings of the International AAAI Conference on Web and
  Social Media (ICWSM)}.

\bibitem[{Mitchell et~al.(2019)Mitchell, Wu, Zaldivar, Barnes, Vasserman,
  Hutchinson, Spitzer, Raji, and Gebru}]{mitchell2019model}
Margaret Mitchell, Simone Wu, Andrew Zaldivar, Parker Barnes, Lucy Vasserman,
  Ben Hutchinson, Elena Spitzer, Inioluwa~Deborah Raji, and Timnit Gebru. 2019.
\newblock Model cards for model reporting.
\newblock In \emph{Proceedings of the Conference on Fairness, Accountability,
  and Transparency}.

\bibitem[{Munger(2017)}]{munger2017tweetment}
Kevin Munger. 2017.
\newblock Tweetment effects on the tweeted: Experimentally reducing racist
  harassment.
\newblock \emph{Political Behavior}, 39(3):629--649.

\bibitem[{Nockleby(2000)}]{nockleby2000hate}
John~T Nockleby. 2000.
\newblock Hate speech.
\newblock \emph{Encyclopedia of the American Constitution}, 3(2):1277--1279.

\bibitem[{O’connell et~al.(1999)O’connell, Pepler, and Craig}]{o1999peer}
Paul O’connell, Debra Pepler, and Wendy Craig. 1999.
\newblock Peer involvement in bullying: Insights and challenges for
  intervention.
\newblock \emph{Journal of Adolescence}, 22(4):437--452.

\bibitem[{Papineni et~al.(2002)Papineni, Roukos, Ward, and
  Zhu}]{papineni2002bleu}
Kishore Papineni, Salim Roukos, Todd Ward, and Wei-Jing Zhu. 2002.
\newblock Bleu: a method for automatic evaluation of machine translation.
\newblock In \emph{Association for Computational Linguistics (ACL)}.

\bibitem[{Qian et~al.(2019)Qian, Bethke, Liu, Belding, and
  Wang}]{qian2019benchmark}
Jing Qian, Anna Bethke, Yinyin Liu, Elizabeth Belding, and William~Yang Wang.
  2019.
\newblock A benchmark dataset for learning to intervene in online hate speech.
\newblock In \emph{Proceedings of EMNLP-IJCNLP}.

\bibitem[{Radford et~al.(2018)Radford, Narasimhan, Salimans, and
  Sutskever}]{radford2018improving}
Alec Radford, Karthik Narasimhan, Tim Salimans, and Ilya Sutskever. 2018.
\newblock \href
  {http://openai-assets.s3.amazonaws.com/research-covers/language-unsupervised/language_understanding_paper.pdf}
  {Improving language understanding by generative pre-training}.

\bibitem[{Ranzato et~al.(2016)Ranzato, Chopra, Auli, and
  Zaremba}]{ranzato2016sequence}
Marc'Aurelio Ranzato, Sumit Chopra, Michael Auli, and Wojciech Zaremba. 2016.
\newblock Sequence level training with recurrent neural networks.
\newblock In \emph{International Conference on Learning Representations
  (ICLR)}.

\bibitem[{de~los Riscos and D’Haro(2020)}]{de2020toxicbot}
Agust{\'\i}n~Manuel de~los Riscos and Luis~Fernando D’Haro. 2020.
\newblock Toxicbot: A conversational agent to fight online hate speech.
\newblock \emph{Conversational Dialogue Systems for the Next Decade}, 704:15.

\bibitem[{Sap et~al.(2020)Sap, Gabriel, Qin, Jurafsky, Smith, and
  Choi}]{sap2020social}
Maarten Sap, Saadia Gabriel, Lianhui Qin, Dan Jurafsky, Noah~A Smith, and Yejin
  Choi. 2020.
\newblock Social bias frames: Reasoning about social and power implications of
  language.
\newblock In \emph{Proceedings of the Association for Computational Linguistics
  (ACL)}.

\bibitem[{Schieb and Preuss(2016)}]{schieb2016governing}
Carla Schieb and Mike Preuss. 2016.
\newblock Governing hate speech by means of counterspeech on facebook.
\newblock In \emph{66th ICA Annual Conference}.

\bibitem[{See et~al.(2019{\natexlab{a}})See, Pappu, Saxena, Yerukola, and
  Manning}]{see2019massively}
Abigail See, Aneesh Pappu, Rohun Saxena, Akhila Yerukola, and Christopher~D
  Manning. 2019{\natexlab{a}}.
\newblock Do massively pretrained language models make better storytellers?
\newblock In \emph{Proceedings of Computational Natural Language Learning
  (CoNLL)}.

\bibitem[{See et~al.(2019{\natexlab{b}})See, Roller, Kiela, and
  Weston}]{see2019makes}
Abigail See, Stephen Roller, Douwe Kiela, and Jason Weston. 2019{\natexlab{b}}.
\newblock What makes a good conversation? how controllable attributes affect
  human judgments.
\newblock In \emph{Proceedings of the 2019 Conference of the North American
  Chapter of the Association for Computational Linguistics: Human Language
  Technologies (NAACL-HLT)}.

\bibitem[{Serban et~al.(2017)Serban, Sordoni, Lowe, Charlin, Pineau, Courville,
  and Bengio}]{serban2017hierarchical}
Iulian~Vlad Serban, Alessandro Sordoni, Ryan Lowe, Laurent Charlin, Joelle
  Pineau, Aaron Courville, and Yoshua Bengio. 2017.
\newblock A hierarchical latent variable encoder-decoder model for generating
  dialogues.
\newblock In \emph{Proceedings of the AAAI Conference on Artificial
  Intelligence (AAAI)}.

\bibitem[{Shao et~al.(2019)Shao, Huang, Wen, Xu et~al.}]{shao2019long}
Zhihong Shao, Minlie Huang, Jiangtao Wen, Wenfei Xu, et~al. 2019.
\newblock Long and diverse text generation with planning-based hierarchical
  variational model.
\newblock In \emph{Proceedings of EMNLP-IJCNLP}.

\bibitem[{Shetty et~al.(2017)Shetty, Rohrbach, Anne~Hendricks, Fritz, and
  Schiele}]{shetty2017speaking}
Rakshith Shetty, Marcus Rohrbach, Lisa Anne~Hendricks, Mario Fritz, and Bernt
  Schiele. 2017.
\newblock Speaking the same language: Matching machine to human captions by
  adversarial training.
\newblock In \emph{Proceedings of the IEEE International Conference on Computer
  Vision (ICCV)}.

\bibitem[{Shi et~al.(2018)Shi, Chen, Qiu, and Huang}]{shi2018toward}
Zhan Shi, Xinchi Chen, Xipeng Qiu, and Xuanjing Huang. 2018.
\newblock Toward diverse text generation with inverse reinforcement learning.
\newblock In \emph{Proceedings of the International Joint Conference on
  Artificial Intelligence (IJCAI)}.

\bibitem[{Shin and Kim(2018)}]{shin2018data}
Youngsoo Shin and Jinwoo Kim. 2018.
\newblock Data-centered persuasion: Nudging user's prosocial behavior and
  designing social innovation.
\newblock \emph{Computers in Human Behavior}, 80:168--178.

\bibitem[{Siegel(2019)}]{siegel2018online}
Alexandra~A Siegel. 2019.
\newblock Online hate speech.
\newblock \emph{Social Media and Democracy}, page~56.

\bibitem[{Sordoni et~al.(2015)Sordoni, Galley, Auli, Brockett, Ji, Mitchell,
  Nie, Gao, and Dolan}]{sordoni2015neural}
Alessandro Sordoni, Michel Galley, Michael Auli, Chris Brockett, Yangfeng Ji,
  Margaret Mitchell, Jian-Yun Nie, Jianfeng Gao, and Bill Dolan. 2015.
\newblock A neural network approach to context-sensitive generation of
  conversational responses.
\newblock In \emph{Proceedings of NAACL-HLT}.

\bibitem[{Srivastava et~al.(2015)Srivastava, Greff, and
  Schmidhuber}]{srivastava2015training}
Rupesh~K Srivastava, Klaus Greff, and J{\"u}rgen Schmidhuber. 2015.
\newblock Training very deep networks.
\newblock In \emph{Advances in Neural Information Processing Systems (NIPS)}.

\bibitem[{Strossen(2018)}]{strossen2018hate}
Nadine Strossen. 2018.
\newblock \emph{Hate: Why we should resist it with free speech, not
  censorship}.
\newblock Oxford University Press.

\bibitem[{Su et~al.(2020)Su, Shen, Zhao, Zhou, Hu, Zhong, Niu, and
  Zhou}]{su2020diversifying}
Hui Su, Xiaoyu Shen, Sanqiang Zhao, Xiao Zhou, Pengwei Hu, Randy Zhong, Cheng
  Niu, and Jie Zhou. 2020.
\newblock Diversifying dialogue generation with non-conversational text.
\newblock In \emph{Proceedings of the Association for Computational Linguistics
  (ACL)}.

\bibitem[{Sutskever et~al.(2014)Sutskever, Vinyals, and
  Le}]{sutskever2014sequence}
Ilya Sutskever, Oriol Vinyals, and Quoc~V Le. 2014.
\newblock Sequence to sequence learning with neural networks.
\newblock In \emph{Advances in Neural Information Processing Systems (NIPS)}.

\bibitem[{Tao et~al.(2018)Tao, Gao, Shang, Wu, Zhao, and Yan}]{tao2018get}
Chongyang Tao, Shen Gao, Mingyue Shang, Wei Wu, Dongyan Zhao, and Rui Yan.
  2018.
\newblock Get the point of my utterance! learning towards effective responses
  with multi-head attention mechanism.
\newblock In \emph{Proceedings of the International Joint Conference on
  Artificial Intelligence (IJCAI)}.

\bibitem[{Tekiroglu et~al.(2020)Tekiroglu, Chung, and
  Guerini}]{tekiroglu2020generating}
Serra~Sinem Tekiroglu, Yi-Ling Chung, and Marco Guerini. 2020.
\newblock Generating counter narratives against online hate speech: Data and
  strategies.
\newblock In \emph{Proceedings of ACL}.

\bibitem[{Tu et~al.(2019)Tu, Ding, Yu, and Gimpel}]{tu2019generating}
Lifu Tu, Xiaoan Ding, Dong Yu, and Kevin Gimpel. 2019.
\newblock Generating diverse story continuations with controllable semantics.
\newblock In \emph{Proceedings of the 3rd Workshop on Neural Generation and
  Translation}.

\bibitem[{Ward(1997)}]{ward1997free}
Kenneth~D Ward. 1997.
\newblock Free speech and the development of liberal virtues: An examination of
  the controversies involving flag-burning and hate speech.
\newblock \emph{U. Miami L. Rev.}, 52:733.

\bibitem[{Warstadt et~al.(2018)Warstadt, Singh, and
  Bowman}]{warstadt2018neural}
Alex Warstadt, Amanpreet Singh, and Samuel~R Bowman. 2018.
\newblock Neural network acceptability judgments.
\newblock \emph{arXiv preprint arXiv:1805.12471}.

\bibitem[{Weingartner and Stahel(2019)}]{weingartner2019online}
Sebastian Weingartner and Lea Stahel. 2019.
\newblock Online aggression from a sociological perspective: An integrative
  view on determinants and possible countermeasures.
\newblock In \emph{Proceedings of the Third Workshop on Abusive Language
  Online}.

\bibitem[{Williams(2019)}]{williams2019hatred}
Matthew Williams. 2019.
\newblock Hatred behind the screens: A report on the rise of online hate
  speech.

\bibitem[{Wu et~al.(2020)Wu, Li, Zhang, Zhou, and Wu}]{wu2020diverse}
Sixing Wu, Ying Li, Dawei Zhang, Yang Zhou, and Zhonghai Wu. 2020.
\newblock Diverse and informative dialogue generation with context-specific
  commonsense knowledge awareness.
\newblock In \emph{Proceedings of the Association for Computational Linguistics
  (ACL)}.

\bibitem[{Xu et~al.(2018)Xu, Ren, Lin, and Sun}]{xu2018diversity}
Jingjing Xu, Xuancheng Ren, Junyang Lin, and Xu~Sun. 2018.
\newblock Diversity-promoting gan: A cross-entropy based generative adversarial
  network for diversified text generation.
\newblock In \emph{Proceedings of the 2018 Conference on Empirical Methods in
  Natural Language Processing (EMNLP)}.

\bibitem[{Zhang et~al.(2020{\natexlab{a}})Zhang, Kishore, Wu, Weinberger, and
  Artzi}]{zhang2020bertscore}
Tianyi Zhang, Varsha Kishore, Felix Wu, Kilian~Q. Weinberger, and Yoav Artzi.
  2020{\natexlab{a}}.
\newblock {BERTScore}: Evaluating text generation with bert.
\newblock In \emph{International Conference on Learning Representations
  (ICLR)}.

\bibitem[{Zhang et~al.(2018)Zhang, Galley, Gao, Gan, Li, Brockett, and
  Dolan}]{zhang2018generating}
Yizhe Zhang, Michel Galley, Jianfeng Gao, Zhe Gan, Xiujun Li, Chris Brockett,
  and Bill Dolan. 2018.
\newblock Generating informative and diverse conversational responses via
  adversarial information maximization.
\newblock In \emph{Advances in Neural Information Processing Systems (NIPS)}.

\bibitem[{Zhang et~al.(2020{\natexlab{b}})Zhang, Sun, Galley, Chen, Brockett,
  Gao, Gao, Liu, and Dolan}]{zhang2020dialogpt}
Yizhe Zhang, Siqi Sun, Michel Galley, Yen-Chun Chen, Chris Brockett, Xiang Gao,
  Jianfeng Gao, Jingjing Liu, and William~B Dolan. 2020{\natexlab{b}}.
\newblock Dialogpt: Large-scale generative pre-training for conversational
  response generation.
\newblock In \emph{Proceedings of the Association for Computational Linguistics
  (ACL)}.

\bibitem[{Zhao et~al.(2017)Zhao, Zhao, and Eskenazi}]{zhao2017learning}
Tiancheng Zhao, Ran Zhao, and Maxine Eskenazi. 2017.
\newblock Learning discourse-level diversity for neural dialog models using
  conditional variational autoencoders.
\newblock In \emph{Proceedings of the Association for Computational Linguistics
  (ACL)}.

\bibitem[{Zhao et~al.(2019)Zhao, Peyrard, Liu, Gao, Meyer, and
  Eger}]{zhao2019moverscore}
Wei Zhao, Maxime Peyrard, Fei Liu, Yang Gao, Christian~M Meyer, and Steffen
  Eger. 2019.
\newblock Moverscore: Text generation evaluating with contextualized embeddings
  and earth mover distance.
\newblock In \emph{Empirical Methods in Natural Language Processing and
  International Joint Conference on Natural Language Processing
  (EMNLP-IJCNLP)}.

\bibitem[{Zhu and Bhat(2020)}]{zhu2020gruen}
Wanzheng Zhu and Suma Bhat. 2020.
\newblock {GRUEN} for evaluating linguistic quality of generated text.
\newblock In \emph{Empirical Methods in Natural Language Processing: Findings
  (Findings of EMNLP)}.

\bibitem[{Zhu et~al.(2021)Zhu, Gong, Bansal, Weinberg, Christin, Fanti, and
  Bhat}]{zhu2021selfsupervised}
Wanzheng Zhu, Hongyu Gong, Rohan Bansal, Zachary Weinberg, Nicolas Christin,
  Giulia Fanti, and Suma Bhat. 2021.
\newblock Self-supervised euphemism detection and identification for content
  moderation.
\newblock In \emph{42nd IEEE Symposium on Security and Privacy}.

\bibitem[{Zhu et~al.(2018)Zhu, Lu, Zheng, Guo, Zhang, Wang, and
  Yu}]{zhu2018texygen}
Yaoming Zhu, Sidi Lu, Lei Zheng, Jiaxian Guo, Weinan Zhang, Jun Wang, and Yong
  Yu. 2018.
\newblock Texygen: A benchmarking platform for text generation models.
\newblock In \emph{The 41st International ACM SIGIR Conference on Research \&
  Development in Information Retrieval}.

\bibitem[{Zhu et~al.(2015)Zhu, Kiros, Zemel, Salakhutdinov, Urtasun, Torralba,
  and Fidler}]{zhu2015aligning}
Yukun Zhu, Ryan Kiros, Rich Zemel, Ruslan Salakhutdinov, Raquel Urtasun,
  Antonio Torralba, and Sanja Fidler. 2015.
\newblock Aligning books and movies: Towards story-like visual explanations by
  watching movies and reading books.
\newblock In \emph{Proceedings of the IEEE International Conference on Computer
  Vision (ICCV)}.

\end{thebibliography}

\clearpage
\appendix

\section{Appendix}

\subsection{Selection of Automatic Metrics}
\label{sec:metric_diversity}

\subsubsection{Diversity}
We measure distinct n-grams (Dist-n) \cite{li2016diversity}, Entropy (Ent-n) \cite{zhang2018generating} and Self-BLEU \cite{zhu2018texygen} for diversity. 

Dist-n reflects the vocabulary diversity by simply dividing the number of unique n-grams by the total number of n-grams of model output. 
One limitation of Dist-n is that it fails to accommodate the frequency difference of n-grams. 
To accommodate the frequency difference of n-grams, we also use the Entropy metric \cite{zhang2018generating}, which reflects how evenly the empirical n-gram distribution is. 

Though Dist-n and Ent-n evaluate the vocabulary diversity well, they fail to evaluate the inter-response diversity. 
For instance, they favor responses with diverse n-grams even when they are highly similar with the rest of the responses.
Therefore, to accommodate such inter-response diversity, we resort to use Self-BLEU \cite{zhu2018texygen} to evaluate how one response resembles the rest in a generated collection of responses. 
Self-BLEU regards one generated sentence as the hypothesis and the other generated sentences as the reference, and calculates the BLEU score for every generated sentence. 
Therefore, the smaller the Self-BLEU, the better the diversity. 


\subsubsection{Relevance}
Most existing works measure relevance \textit{implicitly} by BLEU and ROUGE, a set of metrics evaluating syntactic similarity between the ground truth and the generated output. 
They assume that the ground truth is highly relevant to the conversational input (\ie, it refers to the hate speech in our task) and therefore, the ``closer'' the generated output is to the ground truth, the more relevant the output is to the hate speech instance. 

Explicit relevance evaluation (\ie, relatedness between the conversational input and the generated output) has been studied in only a few existing works. 
For instance, \citet{see2019makes} and \citet{zhang2020dialogpt} ask human annotators to evaluate relevance explicitly. 
\citet{li2020relevance} propose to use HIT-Q and HIT-R, two hit rate based metrics which require hand-crafted rules. 
For automatic metrics, \citet{gao2019jointly} propose to use ``Precision'' to measure relevance. 
However, we consider ``Precision'' inappropriate in our problem setting, because it only measures the relationship between the generated output and the ground truth, but not the relationship between the generated output and the conversational input. 

Since there is no consensus on which automatic metric best serves the purpose of explicit relevance, we select BM25 \cite{manning2008introduction} --- a relevance estimation function widely used in information retrieval. 
Besides, we follow existing works to evaluate implicit relevance by measuring BLEU and ROUGE for syntactic similarity, and MoverScore and BERTScore for semantic similarity.

\subsubsection{Language Quality}
GRUEN \cite{zhu2020gruen} is the only existing open-source unsupervised metric that measures the language quality of generated text. 
It requires no reference to compare with and has been shown to correlate well with human annotations on a variety of language generation tasks.

\subsection{Relevance and Diversity \textit{vs.} Number of Epochs}
\begin{figure}[ht]
	\centering
	\includegraphics[width=0.48\linewidth]{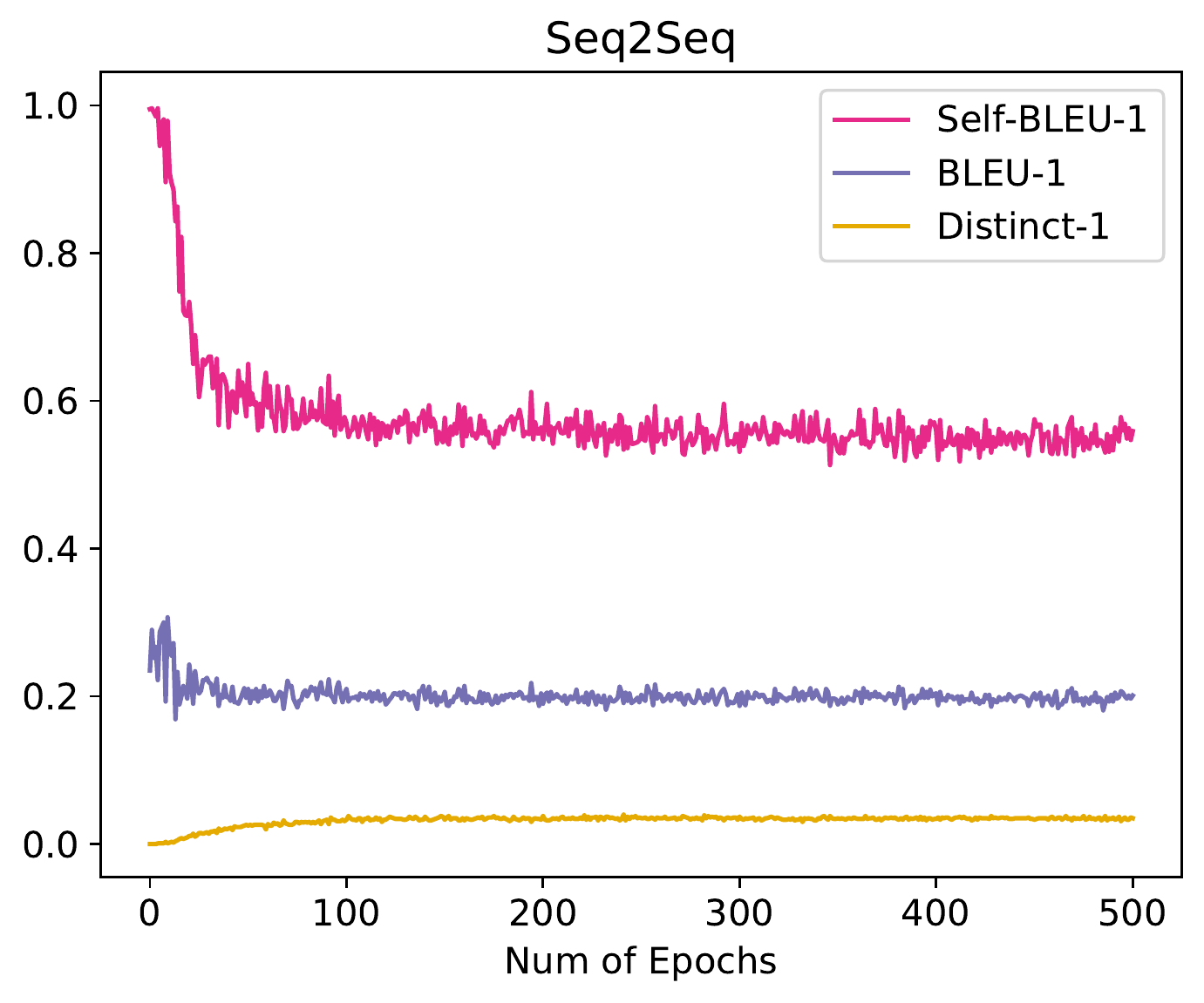}
	\includegraphics[width=0.48\linewidth]{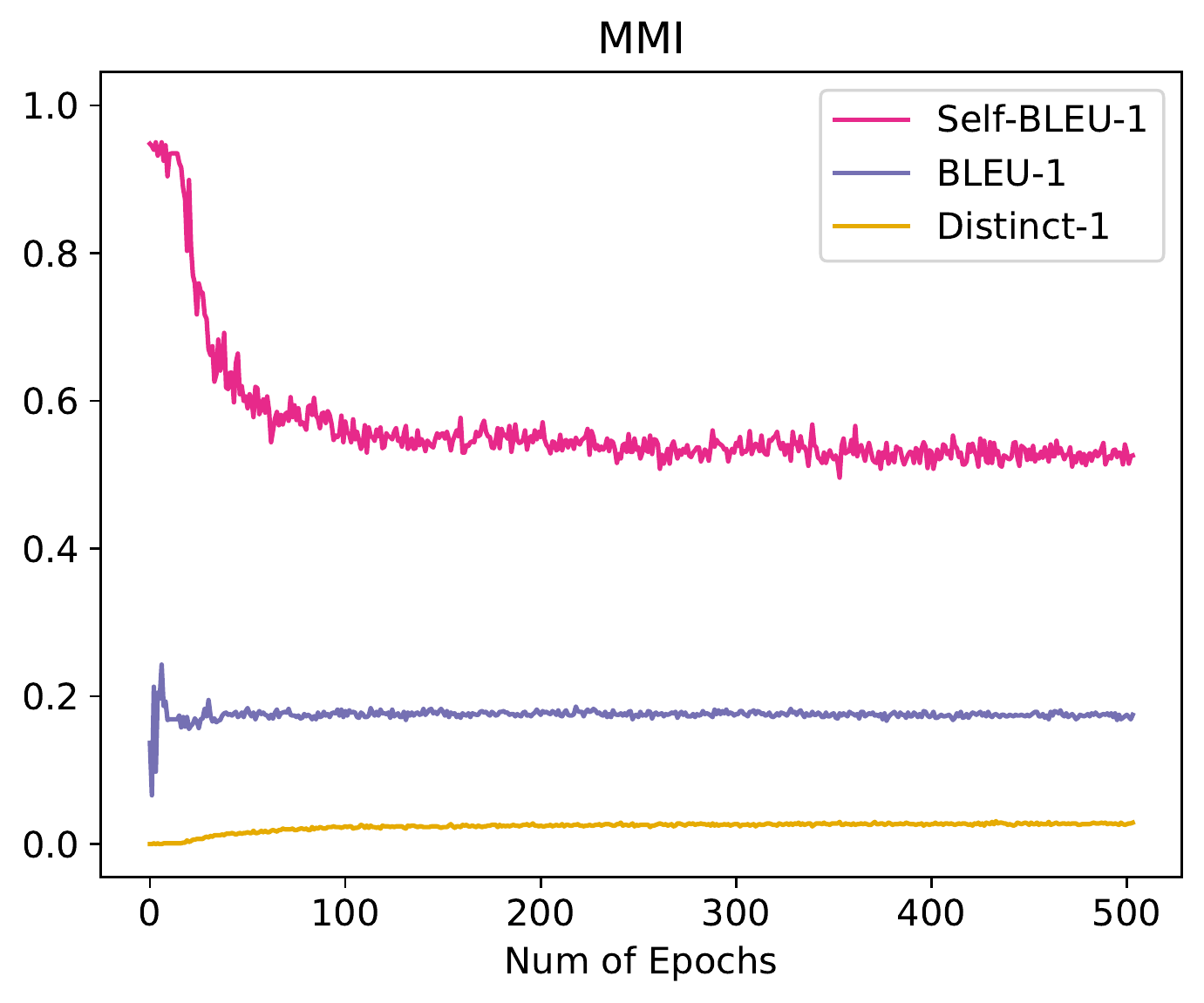}
	\caption{Effects of number of epochs for the Seq2Seq and MMI models on the Gab dataset.}
	\label{fig:diversity_vs_epoch}
\end{figure}

In order to see how the robustness of baseline neural models changes with the number of epochs, 
we plot relevance and diversity measured by automatic metrics against the number of epochs for Seq2Seq and MMI in Figure \ref{fig:diversity_vs_epoch}. 
For each sub-figure, the middle line indicates relevance while the other two lines indicate diversity. 

We note that the diversity increases with the  number of epochs, until  converges at about 100 epochs. 
Surprisingly, the relevance has a spike in the initial few training epochs and then converges to a lower score at about 50 epochs. 
We inspected the results where the spike occurs and observed that the model learns to produce only a few general repetitive counterspeech (\eg, ``Hi there, please refrain from using derogatory comments in the thread. They are hurtful and unwanted. If you continue, Admin will be alerted.") to all hate speech. 
Such general responses, though result in high relevance scores (\eg, BLEU and ROUGE), are not yet effective due to the lack of diversity.
With more training epochs, the models learn to produce more diverse responses at the cost of reduced BLEU and ROUGE. 

Note that all results in this paper (\eg, Table \ref{table:results}) are reported when both relevance and diversity stabilize (\ie, 100 epochs of training). 
\citet{qian2019benchmark} report higher BLEU and ROUGE scores than our results in Table \ref{table:results} for the Seq2Seq model and we suspect that their reported results were obtained with only a few epochs of training.

\subsection{Efficiency Comparison}
We implemented all models in Python 3.7 and conducted all the experiments on a computer with twenty 2.9 GHz Intel Core i7 CPUs and one GeForce GTX 1080 Ti GPU. 
We report the average training time on three datasets. 
Seq2seq: 4.2 hours; 
MMI: 7.8 hours; 
BART: 7.1 hours; 
SpaceFusion: 16.2 hours (running on the CPUs only); 
\our: 4.0 hours. 
We observe that our model requires lower or similar training cost, compared to the baselines. 

\subsection{Case Study}
\label{sec:app_case_study}

Table \ref{table:more_case_study} presents case studies on the generated response for different models. 
In cases (a) and (b), both BART and our model make reasonable responses, whereas Seq2Seq and MMI produce only nonsense. 
In cases (c)-(e), Seq2Seq, MMI and BART generate general and safe responses while our model directly targets the bad words (\eg, ``twat", ``fairy gay faggot") in the hate speech, and even shows understanding and kind warnings to the hate speaker in case (c). 
Therefore, our model may make the hate speaker feel their voices have been heard and is considered closer to human-like moderators. 
Moreover, we find BART sometimes identifies wrong hate words (In case (d), the hate word is ``twat" while BART refers to ``troll" in the response. In case (e), the hate word is ``fairy gay faggot" while BART refers to ``kike".).
The incorrect referral could potentially make the hate speakers irritated and become even more offensive.

\begin{table*}
	\small
	\centering
	\begin{subtable}{1.0\textwidth}
		\begin{tabular}{cc}
			\toprule
			\textbf{Hate Speech} & \multicolumn{1}{p{0.82\textwidth}}{What is the point in listening to Muslim women? They are all retarded.}\\
			\midrule
			\textbf{Seq2Seq}  & \multicolumn{1}{p{0.82\textwidth}}{``So is thinly veiled Muslim Muslim women , the , population , perhaps we go to"}  \\
			\midrule
			\textbf{MMI} & \multicolumn{1}{p{0.82\textwidth}}{``That is influential figures of Baroness Sayeeda Warsi , Malala Yousafzai- I Harvard Fellow Hayfa Al-Kaylani ? Should I list some more brilliantly smart Muslim women ?"} \\
			\midrule
			\textbf{BART} & \multicolumn{1}{p{0.82\textwidth}}{And when CVs submitted under a non Muslim name are 3 times more likely to get an interview than those with a Muslim name attached, the problem very clearly lies with those employers who discriminate against Muslim women who do not want to work with them and instead discriminate against those who do.} \\
			\midrule
			\textbf{\our} & \multicolumn{1}{p{0.82\textwidth}}{The point of listening to Muslim women, and women in general, is that they could broaden and enlighten your view of the world.} \\
			\midrule
		\end{tabular}
		\vspace{-0.2cm}
		\caption{Islamophobia}
		\vspace{+0.3cm}
	\end{subtable}
	
	\begin{subtable}{1.0\textwidth}
		\begin{tabular}{cc}
			\toprule
			\textbf{Hate Speech} & \multicolumn{1}{p{0.82\textwidth}}{All the Muslims should have been punished for yesterday, not only a few. Citizen will take action if the government does not do anything.} \\
			\midrule
			\textbf{Seq2Seq} & \multicolumn{1}{p{0.82\textwidth}}{Are you suggesting that the government deports the good in the countries.} \\
			\midrule
			\textbf{MMI} & \multicolumn{1}{p{0.82\textwidth}}{I am not think the government 's Witness , I is not think that they are not.} \\
			\midrule
			\textbf{BART} & \multicolumn{1}{p{0.82\textwidth}}{What happened to the good old British values of stuff like `tolerance' and `respect for the rule of law' - what you are suggesting sounds like mob rule, vigilante action and people running around with burning torches? If you really care about stopping crime, I think you should focus your energy on campaigning against it, not using it to support your anti-Muslim rhetoric.} \\
			\midrule
			\textbf{\our} & \multicolumn{1}{p{0.82\textwidth}}{You cannot blame all people for actions of a few. Banning something altogether will not solve anything.} \\
			\midrule
		\end{tabular}
		\vspace{-0.2cm}
		\caption{Islamophobia}
		\vspace{+0.3cm}
	\end{subtable}
	
	\begin{subtable}{1.0\textwidth}
		\begin{tabular}{cc}
			\toprule
			\textbf{Hate Speech} & \multicolumn{1}{p{0.82\textwidth}}{I WILL NOT support this one. Not interested in a SJW nigger loving virtue signaler. Will oppose her to all my friends and elected officials.}\\
			\midrule
			\textbf{Seq2Seq}  & \multicolumn{1}{p{0.82\textwidth}}{It does n't make sense to hate an entire race or group of people based on the actions of a few people .}  \\
			\midrule
			\textbf{MMI} & \multicolumn{1}{p{0.82\textwidth}}{We does n't make sense to hate an entire race or religion based on the actions of a few people .} \\
			\midrule
			\textbf{BART} & \multicolumn{1}{p{0.82\textwidth}}{Whoa there, Racist views are hurtful and derogatory. Please be considerate of others when posting. Please refrain from using this type of language. Thank you and have a great day, thank you. (bloody lot of you) (n-word)} \\
			\midrule
			\textbf{\our} & \multicolumn{1}{p{0.82\textwidth}}{I understand you don't care for her, but using words like that will only turn people against you, not her.} \\
			\midrule
		\end{tabular}
		\vspace{-0.2cm}
		\caption{Derogatory Language}
		\vspace{+0.3cm}
	\end{subtable}
	
	\begin{subtable}{1.0\textwidth}
		\begin{tabular}{cc}
			\toprule
			\textbf{Hate Speech} & \multicolumn{1}{p{0.82\textwidth}}{delusional twat}\\
			\midrule
			\textbf{Seq2Seq}  & \multicolumn{1}{p{0.82\textwidth}}{Use of the r-word is unacceptable in discourse as it demeans and insults people with mental disabilities .}  \\
			\midrule
			\textbf{MMI} & \multicolumn{1}{p{0.82\textwidth}}{We can make the same point more effectively without the use of hateful terminology .} \\
			\midrule
			\textbf{BART} & \multicolumn{1}{p{0.82\textwidth}}{Use of the slur here is unacceptable in our discourse as it demeans and insults women, and is hurtful and offensive to the mentally disabled and their families.``Troll" is an offensive term. is an ethnic slur, please use better word choices.} \\
			\midrule
			\textbf{\our} & \multicolumn{1}{p{0.82\textwidth}}{Please refrain using the word twat.} \\
			\midrule
		\end{tabular}
		\vspace{-0.2cm}
		\caption{Derogatory Language}
		\vspace{+0.3cm}
	\end{subtable}
	
	\begin{subtable}{1.0\textwidth}
		\begin{tabular}{cc}
			\toprule
			\textbf{Hate Speech} & \multicolumn{1}{p{0.82\textwidth}}{A political homo? I am not listening to a fairy gay faggot for anyone. Thank you very much.}\\
			\midrule
			\textbf{Seq2Seq}  & \multicolumn{1}{p{0.82\textwidth}}{Stop using hateful language regarding sexual orientation. }  \\
			\midrule
			\textbf{MMI} & \multicolumn{1}{p{0.82\textwidth}}{Please refrain from using hateful words in your posts . It ! It adds nothing of value to the discussion .} \\
			\midrule
			\textbf{BART} & \multicolumn{1}{p{0.82\textwidth}}{Use of the slurs and rank hatred here is unacceptable in our discourse as it demeans and insults and alienates others. ``faggot" and ``kike" are hate speech and are offensive and should be avoided. the word faggot is wrong.} \\
			\midrule
			\textbf{\our} & \multicolumn{1}{p{0.82\textwidth}}{Attacking someone's sexuality (``fairy gay faggot") to express your disagreement with someone's politics is not acceptable in this forum.} \\
			\midrule
		\end{tabular}
		\vspace{-0.2cm}
		\caption{Derogatory Language}
	\end{subtable}
	
	\caption{Examples of the generated counterspeech.}
	\label{table:more_case_study}
\end{table*}

\subsection{Examples of Annotations}
\label{sec:examples_of_annotations}
Table \ref{table:annotation} presents human annotation guidelines and examples on the three dimensions. 
The inter-annotator reliability scores are 0.50, 0.46, 0.36 for diversity, relevance and language quality respectively.

\begin{table*}
	\small
	\centering
	\begin{subtable}{1.0\textwidth}
		\begin{tabular}{|cc}
			\hline
			\multicolumn{1}{|p{0.1\textwidth}}{\rule{0pt}{2ex}\textbf{Hate Speech}} & \multicolumn{1}{p{0.82\textwidth}|}{What is the point in listening to Muslim women? They are all retarded.} \\
			\hline
			\multicolumn{1}{|p{0.1\textwidth}}{\rule{0pt}{2ex}\textbf{System A}} &  \multicolumn{1}{p{0.82\textwidth}|}{``So is thinly veiled Muslim Muslim women , the , population , perhaps we go to"} \\
			\hline
			\multicolumn{1}{|p{0.1\textwidth}}{\rule{0pt}{2ex}\textbf{System B}} & \multicolumn{1}{p{0.82\textwidth}|}{``That is influential figures of Baroness Sayeeda Warsi , Malala Yousafzai- I Harvard Fellow Hayfa Al-Kaylani ? Should I list some more brilliantly smart Muslim women ?"} \\ 
			\hline
			\multicolumn{1}{|p{0.1\textwidth}}{\rule{0pt}{2ex}\textbf{System C}} & \multicolumn{1}{p{0.82\textwidth}|}{And when CVs submitted under a non Muslim name are 3 times more likely to get an interview than those with a Muslim name attached, the problem very clearly lies with those employers who discriminate against Muslim women who do not want to work with them and instead discriminate against those who do.} \\ 
			\hline
			\multicolumn{1}{|p{0.1\textwidth}}{\rule{0pt}{2ex}\textbf{System D}} & \multicolumn{1}{p{0.82\textwidth}|}{The point of listening to Muslim women, and women in general, is that they could broaden and enlighten your view of the world.} \\ 
			\hline 
			\multicolumn{1}{p{0.1\textwidth}}{\rule{0pt}{4ex}\textbf{Question 1:}} & \multicolumn{1}{p{0.82\textwidth}}{Which system-generated counterspeech is most relevant to the hate speech? A relevant counterspeech should address the central aspect of the hateful message and make a coherent conversation. 
			} \\			
			\cmidrule{2-2} 
			\multicolumn{1}{p{0.1\textwidth}}{\rule{0pt}{0ex}\textbf{Question 2:}} & \multicolumn{1}{p{0.82\textwidth}}{Which system-generated counterspeech is most grammaticality correct, readable and fluent?} \\
			\cmidrule{2-2} 
		\end{tabular}
		\vspace{-0.2cm}
		\caption{Evaluation on relevance and language quality}
	\end{subtable}
	\vspace{+0.5cm}
	
	\begin{subtable}{1.0\textwidth}
		\begin{tabular}{|cc}
			\hline
			\multirow{4}{*}{\textbf{System A}}  
			& \multicolumn{1}{p{0.82\textwidth}|}{\rule{0pt}{2ex}1. "Using ``c--ts" to refer to women is offensive, unnecessary and should be avoided.}  \\
			\cline{2-2}
			& \multicolumn{1}{p{0.82\textwidth}|}{\rule{0pt}{2ex}2. Using the foul language will make people more appropriate words.} \\
			\cline{2-2} 
			& \multicolumn{1}{p{0.82\textwidth}|}{\rule{0pt}{2ex}......} \\
			\cline{2-2} 
			& \multicolumn{1}{p{0.82\textwidth}|}{\rule{0pt}{2ex}10. Using the word ``retards" is offensive to the mentally disabled. its a direct attack to their disability.} \\
			\hline
			\multirow{5}{*}{\textbf{System B}} 
			& \multicolumn{1}{p{0.82\textwidth}|}{\rule{0pt}{2ex}1. ``Ret--ds" as used here may offend the mentally disabled and their families; it should be avoided as it adds nothing of substance.} \\
			\cline{2-2} 
			& \multicolumn{1}{p{0.82\textwidth}|}{\rule{0pt}{2ex}2. Please don't use the r-word in your posts. It doesn't help to the discussion in this thread.} \\
			\cline{2-2} 
			& \multicolumn{1}{p{0.82\textwidth}|}{\rule{0pt}{2ex}......} \\
			\cline{2-2} 
			& \multicolumn{1}{p{0.82\textwidth}|}{\rule{0pt}{2ex}10. Please do not use derogatory language for women.} \\
			\hline
			\multirow{9}{*}{\textbf{System C}}  
			& \multicolumn{1}{p{0.82\textwidth}|}{\rule{0pt}{2ex}1. Please refrain from using hateful and ableist language in your posts.  It adds nothing to your argument or the discussion in this thread.  Please refrain in the future if you would like to keep your account active. Thank you, and have a nice day!} \\
			\cline{2-2} 
			& \multicolumn{1}{p{0.82\textwidth}|}{\rule{0pt}{2ex}2. Please refrain from using hateful ableist language in your posts.  It adds nothing productive to the conversation or the sub.  Please refrain from it in the future if you would like to keep your account active. ``Retard" is a hateful word that is used to demean people who struggle with intellectual disability.} \\
			\cline{2-2} 
			& \multicolumn{1}{p{0.82\textwidth}|}{\rule{0pt}{2ex}......} \\
			\cline{2-2} 
			& \multicolumn{1}{p{0.82\textwidth}|}{\rule{0pt}{2ex}10. Using the word ``cunts" is a direct attack against a person based on their gender. Its offensive, unnecessary and should be avoided. ``B--ch" and ``c--t" are hateful terms used to demeans women in a hateful manner.} \\
			\hline
			\multirow{5}{*}{\textbf{System D}}  
			& \multicolumn{1}{p{0.82\textwidth}|}{\rule{0pt}{2ex}1. Feminists are just human beings fighting for their human rights. Please refrain from using the term in a negative context.} \\
			\cline{2-2} 
			& \multicolumn{1}{p{0.82\textwidth}|}{\rule{0pt}{2ex}2. I don't think that you should be spending so much energy defending your right to violence.} \\
			\cline{2-2} 
			& \multicolumn{1}{p{0.82\textwidth}|}{\rule{0pt}{2ex}......} \\
			\cline{2-2} 
			& \multicolumn{1}{p{0.82\textwidth}|}{\rule{0pt}{2ex}10. Right. I cannot stand this either. As a woman I'm annoyed when female characters are forced into the story.} \\
			\hline
			\multicolumn{1}{p{0.1\textwidth}}{\rule{0pt}{4ex}\textbf{Question 3:}}  & \multicolumn{1}{p{0.82\textwidth}}{Which system has the most diversified counterspeech in terms of vocabulary richness, variety in expression and inter-response diversity?} \\
			\cmidrule{2-2} 
		\end{tabular}
		\vspace{-0.2cm}
		\caption{Evaluation on diversity. The hate speech are not shown to the annotators.}
	\end{subtable}
	\caption{Examples of Annotation. We randomize the system outputs to avoid annotators' selection preferences. }
	\label{table:annotation}
\end{table*}

\end{document}